\documentclass[10pt,twocolumn,letterpaper]{article}
\usepackage[pagenumbers]{cvpr}

\usepackage{times}
\usepackage{epsfig}
\usepackage{graphicx}
\usepackage{amsmath}
\usepackage{amssymb}
\usepackage{multirow}
\usepackage{booktabs}

\usepackage{adjustbox}
\usepackage{multirow}
\usepackage{epstopdf}
\usepackage{bbding}
\usepackage{balance}

\usepackage{color, xcolor, colortbl}
\definecolor{LightGray}{gray}{0.9}

\newcommand{\bfsection}[1]{\vspace*{0.1cm}\noindent\textbf{#1.}}

\definecolor{citecolor}{HTML}{229954}
\usepackage[pagebackref=true,breaklinks=true,colorlinks,citecolor=citecolor,bookmarks=false]{hyperref}

    \makeatletter
\def\@fnsymbol#1{\ensuremath{\ifcase#1\or \dagger\or *\or \ddagger\or
   \mathsection\or \mathparagraph\or \|\or **\or \dagger\dagger
   \or \ddagger\ddagger \else\@ctrerr\fi}}
    \makeatother

\newcommand{\red}[1]{\textcolor{red}{#1}}

\begin{document}
\title{Revisiting RCAN: Improved Training for Image Super-Resolution}

\author{%
Zudi Lin$^1$\thanks{Contact email: \texttt{linzudi@g.harvard.edu}} \quad Prateek Garg$^2$\thanks{Works were done during internship at Harvard University.} \quad Atmadeep Banerjee$^{2*}$ \quad Salma Abdel Magid$^1$ \quad Deqing Sun$^3$ \\[1mm] Yulun Zhang$^4$ \quad Luc Van Gool$^4$ \quad Donglai Wei$^{5}$ \quad Hanspeter Pfister$^1$\\[2mm]
$^1$Harvard University\quad $^2$BITS Pilani\quad $^3$Google\quad $^4$ETH Z\"{u}rich\quad $^5$Boston College
}

\maketitle

\begin{abstract}
    Image super-resolution (SR) is a fast-moving field with novel architectures attracting the spotlight. However, most SR models were optimized with dated training strategies. In this work, we revisit the popular RCAN model and examine the effect of different training options in SR. Surprisingly (or perhaps as expected), we show that RCAN can outperform or match nearly all the CNN-based SR architectures published after RCAN on standard benchmarks with a proper training strategy and minimal architecture change. Besides, although RCAN is a very large SR architecture with more than four hundred convolutional layers, we draw a notable conclusion that underfitting is still the main problem restricting the model capability instead of overfitting. We observe supportive evidence that increasing training iterations clearly improves the model performance while applying regularization techniques generally degrades the predictions. We denote our simply revised RCAN as RCAN-it and recommend practitioners to use it as baselines for future research. Code is publicly available at {\small \url{https://github.com/zudi-lin/rcan-it}}.
\end{abstract}
\section{Introduction}

Image super-resolution (SR) is a basic computer vision task that focuses on reconstructing high-resolution (HR) details from corresponding low-resolution (LR) images. The frontier of image SR research has been significantly pushed forward with deep neural networks~\cite{dai2019second,dai2019imageRDAN,dong2016image,dong2016accelerating,kim2016accurate,ledig2017photo,zhang2018image,zhang2018residual}. For the past several years, the field has put spotlights mostly on novel network architectures. 

However, new architectures optimized with improved training strategies are usually compared with earlier models trained with dated protocols. The importance of training strategies that contribute collaboratively to the performances is rarely explored. Recent works on image recognition~\cite{bello2021revisiting,he2019bag} have demonstrated that without or with minimal change to ``traditional'' neural network architectures like ResNet~\cite{he2016deep}, those models can match or surpass the performance of novel architectures proposed after by just improving the training and regularization techniques. Similar observations were also indicated in optical flow estimation~\cite{sun2019models,sun2021autoflow}. Such evidence motivates us to interrogate the training strategies for ``traditional'' SR architectures for better understanding the sources of empirical gains~\cite{lipton2018troubling} in this field and unlocking their potential. 

\begin{figure}[t]
\centering
\includegraphics[width=0.95\columnwidth]{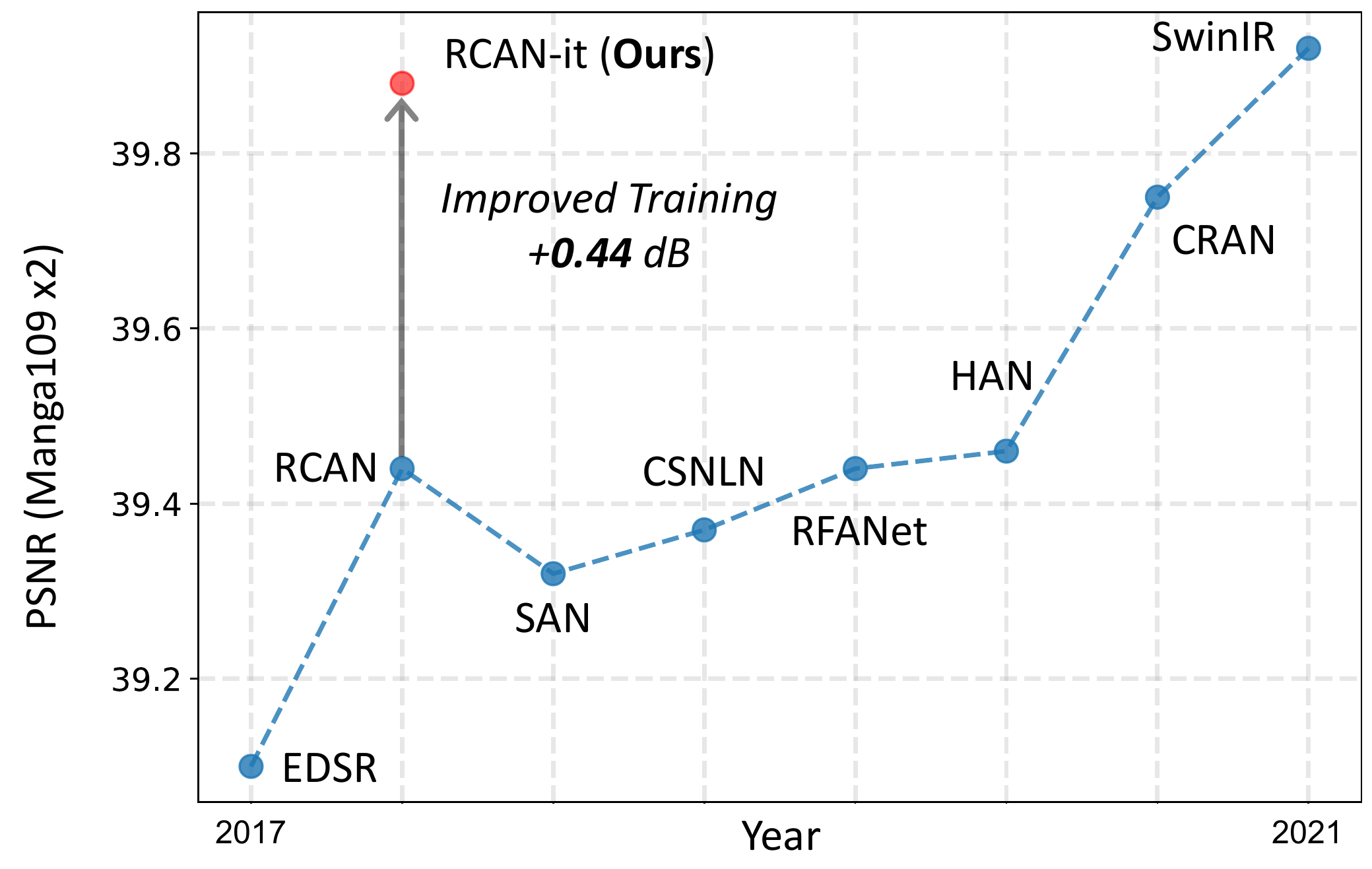}
\caption{
Improving SR with better training. A proper training strategy and minimal architecture change can significantly improve the performance of RCAN~\cite{zhang2018image} and match the empirical gain from novel architectures. Year ($x$-axis) is a rough estimation.
}\label{fig:teaser}
\end{figure}
\begin{table}[t]
\footnotesize
\center
\begin{center}
\caption{Quantitative comparison between original RCAN~\cite{zhang2018image} and our RCAN with improved training (RCAN-it). Results are evaluated by the PSNR (dB) metric for two SR scales.
}

\label{tab:teaser}
\resizebox{\columnwidth}{!}{%
\begin{tabular}{lccccccc}
\toprule
Method & & Set5 & Set14 & B100 & Urban100 & Manga109
\\
\midrule
RCAN~\cite{zhang2018image} & \multirow{2}{*}{x2} 
& 38.27 & 34.12 & 32.41 & 33.34 & 39.44\\
RCAN-it & & {\bf 38.37} & {\bf 34.49} & {\bf 32.48} & {\bf 33.62} & {\bf 39.88}\\
\midrule
RCAN~\cite{zhang2018image} & \multirow{2}{*}{x4} & 32.63 & 28.87 & 27.77 & 26.82 & 31.22\\
RCAN-it & 
& {\bf 32.69}
& {\bf 28.99}
& {\bf 27.87}
& {\bf 27.16}
& {\bf 31.78}\\
\bottomrule
\end{tabular}
} 
\end{center}
\end{table}

To this end, we survey modern training and regularization strategies widely in use for other vision tasks and focus on improving the performance of the Residual Channel Attention Network (RCAN)~\cite{zhang2018image}, which was released more than three years ago and became a popular architecture in the SR field. RCAN is a very deep architecture with more than 400 convolutional layers. However, the potential of this very deep architecture can be hindered by dated training protocols in the original version. Surprisingly (or perhaps as expected according to the observations in image recognition and optical flow estimation), RCAN can outperform or match nearly all subsequent CNN-based approaches that claimed superior performance with a proper training scheme. We denote our improved version of RCAN as RCAN-{\em it} ({\em it} stands for {\em improved training}). We show the comparison with state-of-the-art SR approaches in Figure~\ref{fig:teaser} and the direct improvement upon RCAN paper results in Table~\ref{tab:teaser}. Specifically, our RCAN-it achieves a PSNR of 39.88 dB for $\times2$ SR on Manga109~\cite{matsui2017sketch}, which is an absolute improvement of 0.44 dB over RCAN~\cite{zhang2018image}. The score is better or comparable to the latest works, including CRAN~\cite{Zhang2021ContextRA} and SwinIR~\cite{liang2021swinir}. With self-ensemble inference, the PSNR is further increased to {\bf 40.04} dB, outperforming all existing approaches with or without self-ensemble.


We conduct comprehensive ablation studies from the training perspective and have several notable results. First, although RCAN is a huge SR architecture, we notice that the model performance is currently restricted by {\em underfitting} instead of overfitting when training on the widely used DF2K dataset~\cite{lim2017enhanced}. We draw this conclusion based on the observations that simply increasing the training iterations can clearly improve the prediction scores, while regularization techniques effective for image recognition models generally decrease the SR model performance\footnote{More precisely, those regularization techniques decrease the performance under the hyper-parameters we tested.}. Second, using a large-batch training scheme with state-of-the-art optimizers and learning-rate scheduling rules, the model can already match the results of the original training protocol with $4\times$ less training time. Third, a simple architecture modification that changes ReLU nonlinearity to SiLU~\cite{hendrycks2016gaussian,Ramachandran2017SwishAS} achieves similar improvement as training the baseline model $2\times$ longer. Fourth, with {\em warm-start} that initializes SR networks with different scales using the pretrained $\times2$ model, we can further save the training time and cost for $\times3$ and $\times4$ models while still achieving competitive performance. We also conducted an oracle study on the benchmark sets and showed that there is still large space for improvement before reaching the upper bound of the RCAN architecture, shedding light on future directions like diversifying training data and domain adaptation.

To summarize, our work makes three main contributions. First, different from recent image SR publications that emphasize architectures, we investigate the optimization aspect of this low-level vision task and identify the {\em underfitting} problem in deep SR networks. Second, with an effective large-batch training framework, updated learning protocol, and minimal architectural modification, our RCAN-it outperforms or matches the latest approaches released more than three years later than RCAN~\cite{zhang2018image}. Third, our comprehensive empirical studies demonstrate the interplay of different training strategies and provide practitioners with valuable information for future research.

\section{Related Work}

\bfsection{Image super-resolution}
Image super-resolution is a classic low-level computer vision problem~\cite{freeman2002example}. Deep neural networks have become the de facto methodology for image SR in the past several years thanks to their ability to model the end-to-end mapping between LR and HR images. Starting from the pioneering work where the model is a simple three-layer CNN~\cite{dong2016image}, developing better architectures has then become the central topic in this field. Subsequent improvements are achieved by increasing the depth and width of models~\cite{kim2016accurate}, applying residual~\cite{lim2017enhanced,zhang2018image} and dense~\cite{zhang2018residual} connections, as well as introducing different channel and spatial attention techniques~\cite{zhang2018image,dai2019second,niu2020single,dai2019imageRDAN,magid2021dynamic}. Recent work even gets rid of the CNN-based architectures and utilizes a transformer~\cite{vaswani2017attention} (specifically a Swin Transformer~\cite{liu2021swin}) for image restoration~\cite{liang2021swinir}. We champion the success of better SR architectures, but from a machine learning perspective, good predictions result from the interplay of architecture, training data, and optimization strategies. Previous works clearly demonstrate that image SR can benefit from better architectures originally proposed for high-level vision tasks~\cite{he2016deep,hu2018squeeze,liu2021swin}, but pays less attention to the training strategies that achieve comparable or even large empirical gains than architectures~\cite{he2019bag,bello2021revisiting}. Therefore we extensively investigate the rarely interrogated training strategies in the SR field and show good practice for training  ``traditional'' deep SR architectures like RCAN~\cite{zhang2018image}.

\bfsection{Training and regularization strategies} Image recognition literature has demonstrated that a proper training strategy can effectively decrease the training time using parallelism~\cite{goyal2017accurate,you2017large} and increase the performance with better optimization strategies and regularization techniques~\cite{he2019bag,bello2021revisiting}. In large-batch training, the learning rate needs to be scaled (usually linearly with the batch size~\cite{goyal2017accurate}) to compensate for the reduced number of parameter updates. Some works use warm-up~\cite{goyal2017accurate,he2019bag} to avoid the sudden increase of learning rate for training stability. Optimizers like LAMB~\cite{you2019large} improve upon Adam~\cite{kingma2014adam} with a layer-wise adaptive mechanism. We adopt large-batch training for RCAN~\cite{zhang2018image} in image SR to effectively decrease the training time while achieving comparable performance as the original protocol.

Regularizations like weight decay, stronger data augmentations~\cite{cubuk2020randaugment}, mixup~\cite{zhang2017mixup} and stochastic depth~\cite{huang2016deep} usually boost the accuracy of deep classification models. We thus study their effects on image SR, a basic {\em low-level} vision topic. Different from image recognition where the main challenge is overfitting (\eg, training ResNet~\cite{he2016deep} for more epochs decrease the classification accuracy~\cite{bello2021revisiting}), we show evidence that the performance of deep SR models~\cite{zhang2018image} is restricted by {\em underfitting}, demonstrating the unique characteristics of different computer vision fields. We thus provide a practical training strategy for SR models.
\section{Methodology}

In this section, we first revisit the base RCAN architecture and show a small modification to the activation function~(Sec.~\ref{sec:arch}). We then discuss the training and regularization strategies tested throughout the paper (Sec.~\ref{sec:train}).

\subsection{Architecture}\label{sec:arch}
The Residual Channel Attention  Network  (RCAN)~\cite{zhang2018image} is a popular SR architecture that features three main improvements upon previous work like EDSR~\cite{lim2017enhanced}. First, it uses a squeeze-and-excitation (or channel-attention) block~\cite{hu2018squeeze} after the second $3\times3$ convolution layer in each residual block to re-weight the importance of different channels. Second, it has a novel residual-in-residual design with {\em long} skip connections over multiple residual blocks to bypass low-frequency information and facilitate better learning of high-frequency details. Third, RCAN is a very deep architecture with a large model capacity as it has 200 residual blocks and more than 400 convolutional layers. In this work, we keep the RCAN architecture mostly untouched. The only modification we tested from the architectural perspective is to substitute the original ReLU activation with the  {\em Sigmoid Linear Unit} (SiLU)~\cite{hendrycks2016gaussian} (also known as Swish~\cite{Ramachandran2017SwishAS}) activation function:
\begin{equation}
    f(x) = x \cdot \sigma(x),
\end{equation}
where $\sigma(x)$ is the sigmoid function. Different from ReLU, SiLU is a smooth, non-monotonic function. Previous results~\cite{Ramachandran2017SwishAS} show consistent improvement of SiLU over ReLU on image recognition benchmarks but have not yet explored its impact for low-level vision tasks like image SR.

\subsection{Training Strategies}\label{sec:train}

\bfsection{Large-batch optimization} The original RCAN was trained with the Adam~\cite{kingma2014adam} optimizer, a small batch size (BS) of 16, and a small learning rate ($\eta$) of $10^{-4}$. The consequence is slow convergence for about 7 days on a single GPU device (Table.~\ref{tab:baseline}, 1st row). Therefore our first improvement over the original training protocol is to enable {\em large-batch} training over multiple GPUs for faster convergence. Since the total number of gradient updates decreases, we apply a linear scaling rule so that when we multiply the mini-batch size by $k$, the learning rate is also multiplied by $k$. For training stability, we use Lamb~\cite{you2019large}, a layerwise adaptive optimizer designed for large-batch training. We also substitute the original multi-step learning rate scheduler with cosine annealing~\cite{loshchilov2016sgdr}, whose only hyper-parameter is the total number of iterations (or epochs). We effectively increase the batch size by $\times16$ and largely save training time using parallelism by employing those techniques.

\begin{figure}[t]
\centering
\includegraphics[width=\columnwidth]{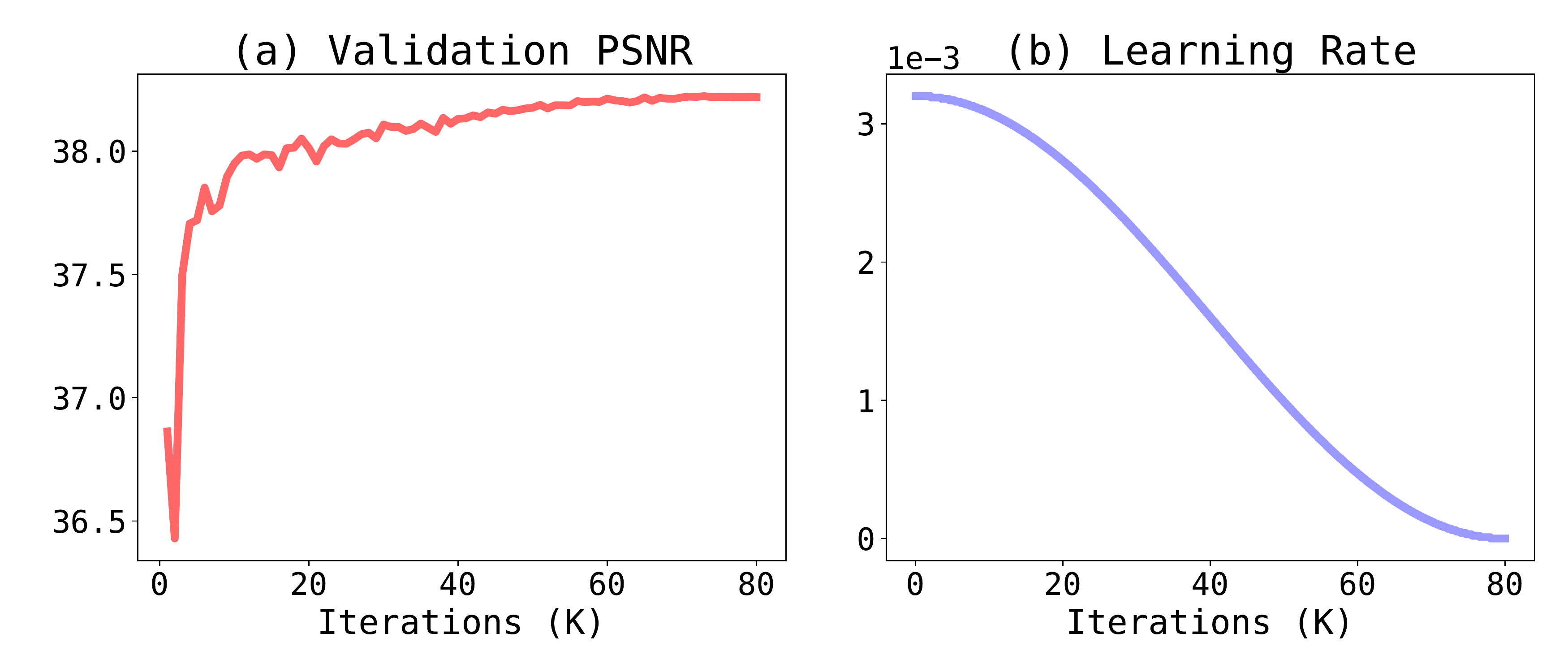}
\caption{Validation curve of baseline RCAN. We show {\bf (a)} $\times2$ PSNR on the DF2K validation set and {\bf (b)} learning rate schedule for the highlighted baseline in Table~\ref{tab:baseline}. The PSNR performance is still increasing by the end of the training, indicating {\em underfitting}.
}\label{fig:validation}
\end{figure}

\bfsection{Longer training}.We observe supportive evidence that the validation performance of RCAN is still increasing by the end of the baseline training protocol (Fig.~\ref{fig:validation}\red{a}), indicating the {\em underfitting} problem. Therefore, we apply a straightforward but effective technique, increasing the number of training iterations, to alleviate the challenge of underfitting. Unlike image classification results where longer training decreases the performance~\cite{bello2021revisiting} due to overfitting, we show in experiments that such a strategy effectively improve the performance of deep SR models like RCAN.

\bfsection{Large-patch finetuning} Previous results suggest that training with larger patches improves the performance~\cite{liang2021swinir}. However, increasing input height and width under the large-batch setting will quadratically increase the GPU memory usage, making the training costly or even infeasible under a hardware budget. Therefore we design a two-stage training strategy where the model is first optimized with standard patch size ($48\times48$) as RCAN~\cite{zhang2018image}. We then finetune the model with larger patches ($64\times64$) for a smaller number of iterations to enjoy the benefit of a larger input {\em field-of-view} (batch size is reduced accordingly to fit GPU memory).  

\bfsection{Low-precision training} In popular deep learning frameworks like PyTorch~\cite{paszke2019pytorch}, neural networks are optimized with 32-bit float point (FP32) precision. Existing work for image recognition indicates that low-precision training (reduces data, model parameters, and gradients to FP16 precision) can preserve or even slightly improve the original model accuracy with significantly decreased training time and GPU memory usage~\cite{he2019bag}. We therefore also examine this technique with RCAN and show that low-precision (FP16) training has different behaviors in image SR.

\bfsection{Regularization techniques} Besides improving training strategies, we also tested regularization techniques widely used for image recognition, including stronger data augmentations, mixup~\cite{zhang2017mixup} and stochastic depth~\cite{huang2016deep}. For augmentations, we add random invert and channel shuffle. For mixup we use $\text{Beta}(0.15, 0.15)$ to produce random interpolation weights. For stochastic depth, we randomly skip a residual block with a probability of 0.5. Although adding regularizations can effectively boost the performance of large image recognition models~\cite{huang2016deep,bello2021revisiting}, we show in experiments that RCAN does not benefit from regularizations as it suffers from {\em underfitting} instead of overfitting (Sec.\ref{sec:exp_train}).

\bfsection{Warm start} For different SR scales, $\times k, k \in \{2,3,4\}$, most RCAN layers have identical structures except the \texttt{tail} module that contains unique convolution and pixel shuffle~\cite{shi2016real} layers. Therefore when having a pretrained $\times2$ model, we can directly transfer the weights to $\times3$ and $\times4$ models instead of training them from scratch, which is denoted as {\em warm start}. Since the \texttt{tail} modules are not shared, we first finetune only the random initialized \texttt{tail} module with other layers frozen until convergence, which takes $<1$ hour. We then finetune the whole model with only $50\%$ of the normal iterations to save time and cost.
\section{Experiments}

We first describe the data, metrics and baseline models (Sec.~\ref{sec:exp_setup}). We then show the impacts of different training strategies (Sec.~\ref{sec:exp_train}) and benchmark comparisons (Sec.~\ref{sec:benchmark}).

\subsection{Setup}\label{sec:exp_setup}

\bfsection{Dataset and metric} 
Following latest publications~\cite{Zhang2021ContextRA,magid2021dynamic,liang2021swinir}, we use the DF2K dataset for training, which combines DIV2K~\cite{timofte2017ntire} and Flickr2K~\cite{lim2017enhanced} together to create a single training set with totally 3,550 images. As a common practice, we evaluate models on five standard benchmark datasets including Set5~\cite{bevilacqua2012low}, Set14~\cite{zeyde2012single}, B100~\cite{martin2001database}, Urban100~\cite{huang2015single}, and Manga109~\cite{matsui2017sketch}. For the metrics, we report the peak signal-to-noise ratio (PSNR) in all experiments, and also the structural similarity metric (SSIM)~\cite{wang2004image} in benchmark comparison. At evaluation time, the RGB images are first transformed into the YCbCr space, and the metrics are applied to the Y channel (\ie, luminance).

\begin{table}[t]
\center
\begin{center}
\caption{Baseline results (PSNR) for $\times2$ SR. We train RCAN\cite{zhang2018image} on DF2K with the original protocol (1st row) and our large-batch training strategies (2nd and 3rd rows). The {\setlength{\fboxsep}{2pt}\colorbox{green!25}{highlighted}} setting matches the original strategy and reduces the training time by $77\%$, which is used as the baseline for further improvements.}

\label{tab:baseline}
\resizebox{0.97\columnwidth}{!}{%
\begin{tabular}{lccccc}
\toprule
BS, $\eta$ & Time & Set5 & Set14 & B100 & Urban100 
\\
\midrule
16, ~~0.0001 & 7 Days & 38.35 & 34.33 & 32.48 & 33.59 \\
256, 0.0016 & 1.6 Days & 38.34 & 34.35 & 32.45 & 33.52 \\
\rowcolor{green!25}
256, 0.0032 & 1.6 Days & 38.35 & 34.42 & 32.46 & 33.61 \\
\bottomrule
\end{tabular}
} 
\end{center}
\end{table}

\bfsection{Updated RCAN Baseline}
We use the standard RCAN model, which consists of ten {\em residual groups} (RG), while each RG contains 20 residual blocks and one convolution layer. The number of channels for all layers is fixed to 64, except the input and upsampling layers. We use input patches of size 48$\times$48 as RCAN. For simplicity and consistency among experiments, we only use the pixel-wise mean-absolute-error ($\ell_1$) loss. Inputs are augmented with random horizontal and vertical flips as well as transpose. 

\begin{table*}[t]
\center
\begin{center}
\caption{
Ablation studies of training options. The impact of {\setlength{\fboxsep}{2pt}\textbf{\colorbox{blue!20}{architecture modifications}}},  {\setlength{\fboxsep}{2pt}\textbf{\colorbox{yellow!30}{training strategies}}} and {\setlength{\fboxsep}{2pt}\textbf{\colorbox{red!20}{regularizations}}} are shown upon the updated RCAN baseline (highlighted in Table~\ref{tab:baseline}) for $\times 2$ SR. $^\dagger$Longer training means $2\times$ the total iterations of baseline. $^\ddagger$We finetune on larger patches instead of training from scratch to save computing resource. Results are produced with self-ensemble.
}
\label{tab:individual}
\resizebox{\textwidth}{!}{%
\begin{tabular}{lcccccccccccc}
\toprule
\multirow{2}{*}{Configuration}
& \multicolumn{2}{c}{Set5} 
& \multicolumn{2}{c}{Set14} 
& \multicolumn{2}{c}{B100} 
& \multicolumn{2}{c}{Urban100} 
& \multicolumn{2}{c}{Manga109}
& \multicolumn{2}{c}{Average}
\\
\cmidrule{2-13}
& PSNR & $\Delta$ 
& PSNR & $\Delta$
& PSNR & $\Delta$
& PSNR & $\Delta$
& PSNR & $\Delta$
& PSNR & $\Delta$
\\
\midrule
Baseline &
38.350 & - &	34.415 & - & 32.463 & -	& 33.610 & - & 39.866 & - & 35.741 & - \\
\cellcolor{blue!20} w/ SiLU Activation
& 38.369 & \textbf{\textcolor{citecolor}{+0.019}}
& 34.463 & \textbf{\textcolor{citecolor}{+0.048}} 
& 32.475 & \textbf{\textcolor{citecolor}{+0.012}} 
& 33.669 & \textbf{\textcolor{citecolor}{+0.059}} 
& 39.936 & \textbf{\textcolor{citecolor}{+0.070}}
& 35.783 & \textbf{\textcolor{citecolor}{+0.042}} \\
\cellcolor{yellow!30} w/ Longer Training$^\dagger$
& 38.395 & \textbf{\textcolor{citecolor}{+0.045}}
& 34.438 & \textbf{\textcolor{citecolor}{+0.023}}
& 32.490 & \textbf{\textcolor{citecolor}{+0.028}}
& 33.656 & \textbf{\textcolor{citecolor}{+0.047}}
& 39.961 & \textbf{\textcolor{citecolor}{+0.095}}
& 35.788 & \textbf{\textcolor{citecolor}{+0.047}} \\
\cellcolor{yellow!30} w/ Larger Patches$^\ddagger$ 
& 38.367 & \textbf{\textcolor{citecolor}{+0.017}}
& 34.473 & \textbf{\textcolor{citecolor}{+0.058}}
& 32.484 & \textbf{\textcolor{citecolor}{+0.021}}
& 33.687 & \textbf{\textcolor{citecolor}{+0.077}}	
& 39.929 & \textbf{\textcolor{citecolor}{+0.063}}
& 35.788 & \textbf{\textcolor{citecolor}{+0.047}} \\
\midrule
\cellcolor{yellow!30} w/ FP16 Precision
& 38.360 & \textbf{\textcolor{citecolor}{+0.010}}
& 34.379 & \textbf{\textcolor{red!60}{-0.036}}
& 32.450 & \textbf{\textcolor{red!60}{-0.013}}
& 33.530 & \textbf{\textcolor{red!60}{-0.080}}
& 39.812 & \textbf{\textcolor{red!60}{-0.054}}
& 35.706 & \textbf{\textcolor{red!60}{-0.035}}
\\
\cellcolor{red!20} w/ Color Augmentation
& 38.341 & \textbf{\textcolor{red!60}{-0.009}}	
& 34.414 & \textbf{\textcolor{red!60}{-0.001}}		
& 32.457 & \textbf{\textcolor{red!60}{-0.006}}		
& 33.592 & \textbf{\textcolor{red!60}{-0.017}}		
& 39.904 & \textbf{\textcolor{citecolor}{+0.038}}
& 35.742 & \textbf{\textcolor{citecolor}{+0.001}} \\
\cellcolor{red!20} w/ Mixup
& 38.354 & \textbf{\textcolor{citecolor}{+0.004}}	
& 34.328 & \textbf{\textcolor{red!60}{-0.087}}		
& 32.453 & \textbf{\textcolor{red!60}{-0.009}}		
& 33.532 & \textbf{\textcolor{red!60}{-0.078}}		
& 39.853 & \textbf{\textcolor{red!60}{-0.013}}
& 35.704 & \textbf{\textcolor{red!60}{-0.037}} \\
\cellcolor{red!20} w/ Stochastic Depth
& 38.184 & \textbf{\textcolor{red!60}{-0.166}}
& 33.902 & \textbf{\textcolor{red!60}{-0.513}}	
& 32.335 & \textbf{\textcolor{red!60}{-0.128}}
& 32.760 & \textbf{\textcolor{red!60}{-0.850}}
& 39.276 & \textbf{\textcolor{red!60}{-0.590}}
& 35.292 & \textbf{\textcolor{red!60}{-0.449}}\\
\bottomrule
\end{tabular}
} 
\end{center}
\end{table*}

\begin{table*}[t]
\center
\begin{center}
\caption{
Additive improvement of training options ($\times 2$ SR PSNR). Upon the updated RCAN baseline (highlighted in Table~\ref{tab:baseline}), combining SiLU and longer training can increase the average score by 0.1 dB. Fine-tuning on larger patches ($64\times 64$) further improves the performance. $^\star$Oracle denotes the model finetuned on the benchmark sets, which indicates the upper bound of RCAN.
}
\label{tab:increment}
\resizebox{\textwidth}{!}{%
\begin{tabular}{lcccccccccccc}
\toprule
\multirow{2}{*}{Configuration}
& \multicolumn{2}{c}{Set5} 
& \multicolumn{2}{c}{Set14} 
& \multicolumn{2}{c}{B100} 
& \multicolumn{2}{c}{Urban100} 
& \multicolumn{2}{c}{Manga109}
& \multicolumn{2}{c}{Average}
\\
\cmidrule{2-13}
& PSNR & $\Delta$ 
& PSNR & $\Delta$
& PSNR & $\Delta$
& PSNR & $\Delta$
& PSNR & $\Delta$
& PSNR & $\Delta$
\\
\midrule
Baseline &
38.350 & - &	34.415 & - & 32.463 & -	& 33.610 & - & 39.866 & - & 35.741 & - \\
{\bf +} SiLU \& Train Longer
& 38.405 & \textbf{\textcolor{citecolor}{+0.055}}
& 34.527 & \textbf{\textcolor{citecolor}{+0.112}} 
& 32.497 & \textbf{\textcolor{citecolor}{+0.034}} 
& 33.747 & \textbf{\textcolor{citecolor}{+0.137}} 
& 40.003 & \textbf{\textcolor{citecolor}{+0.137}}
& 35.836 & \textbf{\textcolor{citecolor}{+0.095}} \\
{\bf +} Larger Patches 
& {\bf 38.406} & \textbf{\textcolor{citecolor}{+0.001}}
& {\bf 34.546} & \textbf{\textcolor{citecolor}{+0.019}}
& {\bf 32.511} & \textbf{\textcolor{citecolor}{+0.014}}
& {\bf 33.790} & \textbf{\textcolor{citecolor}{+0.043}}	
& {\bf 40.041} & \textbf{\textcolor{citecolor}{+0.038}}
& {\bf 35.859} & \textbf{\textcolor{citecolor}{+0.023}} \\
\midrule
Oracle$^\star$
& 39.273 & \textbf{\textcolor{black!35}{+0.867}}
& 35.010 & \textbf{\textcolor{black!35}{+0.464}}
& 32.622 & \textbf{\textcolor{black!35}{+0.111}}
& 34.006 & \textbf{\textcolor{black!35}{+0.216}}
& 40.321 & \textbf{\textcolor{black!35}{+0.280}}
& 36.338 & \textbf{\textcolor{black!35}{+0.388}} \\
\bottomrule
\end{tabular}
} 
\end{center}
\end{table*}

Since inputs are uniformly cropped from large training images, we use {\em iterations} instead of epochs to denote the training length. The original strategy uses a batch size (BS) of 16 and a learning rate ($\eta$) of $10^{-4}$. The model is optimized with ADAM~\cite{kingma2014adam} ($\beta_1$=0.9, $\beta_2$=0.99) for 1,725K iterations, and $\eta$ is halved every $20\%$ of the total iterations. In our large-batch scheme, we use a BS$=256$ and scale $\eta$ with the linear scaling rule. For stability, we change ADAM to the Lamb~\cite{you2019large} optimizer. The original training scheme used a single Nvidia V100 GPU and finished the training in about 7 days. We use 4 V100 GPUs in our large-batch setting and finish the training in 1.6 days for 80K iterations. The system is implemented with PyTorch~\cite{paszke2019pytorch}.

Table~\ref{tab:baseline} shows that if we strictly follow the linear scaling rule to scale the $\eta$ from $10^{-4}$ ($0.0016 = \frac{256}{16} \cdot 10^{-4}$), there is a performance gap to the original protocol. If we scale $\eta$ from $2\cdot10^{-4}$, the performance becomes comparable to the original setting but only uses $23\%$ of the original training time. Therefore, we set BS=256, $\eta$=0.0032 as the baseline for investigating the impact of other training techniques in the following experiments. Note that the original setting results are better than the RCAN paper results~\cite{zhang2018image} as it is optimized on the larger DF2K dataset, while the paper results are produced with DIV2K~\cite{timofte2017ntire} only.

We also observe that training is unstable with large learning rates. When $\eta$=0.0064, the model produces undefined numbers (NaN) after $\sim$2K iterations. For $\eta$=0.0048, the model produces NaN after $\sim$16K iterations. Therefore with BS=256, increasing $\eta$ for faster convergence is not practicable. Techniques like gradient clipping may alleviate the problem, and we leave that for future exploration.

\subsection{Impact of Training Strategies}\label{sec:exp_train}

\bfsection{Individual changes}
We show the ablation studies of architecture modifications, different training strategies, and regularizations in Table~\ref{tab:individual}. Changing ReLU to SiLU, longer training (training RCAN for 160K iterations, $2\times$ the 80K iteration baseline), and finetuning with larger patches ($64\times 64$) can consistently improve the PSNR scores on all five benchmark datasets when working {\em independently} with the baseline. The average improvements for those three updates are at least 0.042 dB. Notably, the observation that longer training and large-patch finetuning improve the test performance further supports our claim that current deep SR models like RCAN suffers from {\em underfitting}, while those strategies are reasonable for alleviating the problem.

Unlike in image classification~\cite{he2019bag}, low-precision (FP16) training and regularization techniques generally degrade the performance (Table~\ref{tab:individual}, bottom half). Although FP16 training can reduce $50\%$ GPU memory and over $30\%$ of training time, it decreases the average PSNR by 0.035 and sometimes outputs NaN even with the baseline learning rate.  Stochastic depth, a strong regularization that works well for training very deep classification models~\cite{huang2016deep,bello2021revisiting}, decreases the SR scores significantly. We argue this is because randomly silencing residual blocks in a dense {\em regression} task makes a relatively large change in the loss, causing inaccurate gradient estimation. One exception in regularizations is that since Manga109~\cite{matsui2017sketch} consists of images from manga (comic books) while the training set contains mostly natural images, color augmentations have a positive effect on this benchmark. In summary, the results show that techniques with empirical gains in image recognition can have different behaviors in image super-resolution, demonstrating the uniqueness of computer vision domains. We thus suggest SR researchers be careful with regularizations when training models on relatively large datasets like DF2K.

\bfsection{Additive improvement} Based on the individual results (Table~\ref{tab:individual}), we keep techniques that consistently improve model performance into our protocol and study their additive effect (Table~\ref{tab:increment}). Longer training of the SiLU model boosts the average PSNR by 0.095 dB, even better than adding up the individual improvements of both techniques. Large-patch finetuning has a smaller effect compared with starting from baseline but still consistently increase the scores and achieve a state-of-the-art PSNR of {\bf 40.04} on Manga109. So far, the best model is optimized for 200K iterations (160K training $+$ 40K large-patch finetuning) on 4 V100 GPUs for 4 days. We believe additional training can further improve the scores (\eg, SwinIR~\cite{liang2021swinir} is optimized for more than 500K iterations).

We also show the {\em oracle} results to understand the upper bound of the RCAN architecture (Table~\ref{tab:increment}, last line). The best $\times2$ model is finetuned on each test set independently until convergence. Interestingly, the performance on the two smallest benchmark sets (\ie, Set5 and Set14) is farthest from saturation. The gap between the best model and oracle (an average PSNR of 0.388 dB) indicates that we still have large space for improvement even with the relatively ``dated'' RCAN architecture. Besides, Set5~\cite{bevilacqua2012low} is known to have JPEG artifacts, while Manga109~\cite{matsui2017sketch} consists of comic images whose statistics significantly differ from the training data. Therefore we suggest future research to focus on diversifying training data, designing better optimization strategies, and testing domain adaptation to further narrow the performance gap besides upgrading the model.

\begin{table*}[t]
\center
\begin{center}
\caption{
Effect of finetuning from {\em warm start} for $\times3$ and $\times4$ SR. By finetuning only the \texttt{tail} module from a pretrained $\times2$ architecture, the model can already match RCAN paper results~\cite{zhang2018image}. Normal finetuning for 80K iterations ($50\%$ of the longer training settings) significantly boosts the performance. Large-patch finetuning still increases the scores, but the improvement is relatively marginal.
}
\label{tab:warm_start}
\resizebox{\textwidth}{!}{%
\begin{tabular}{lcccccccccccc}
\toprule
\multirow{2}{*}{Configuration}
& \multicolumn{2}{c}{Set5} 
& \multicolumn{2}{c}{Set14} 
& \multicolumn{2}{c}{B100} 
& \multicolumn{2}{c}{Urban100} 
& \multicolumn{2}{c}{Manga109}
& \multicolumn{2}{c}{Average}
\\
\cmidrule{2-13}
& PSNR & $\Delta$ 
& PSNR & $\Delta$
& PSNR & $\Delta$
& PSNR & $\Delta$
& PSNR & $\Delta$
& PSNR & $\Delta$
\\
\midrule
Warm Start ($\times\textbf{3}$ SR) 
& 34.878 & - & 30.749 & - & 29.377 & - & 29.237 & - & 34.883 & - & 31.825 & - \\
{\bf +} Normal Finetuning
& 34.935 & \textbf{\textcolor{citecolor}{+0.057}}
& 30.847 & \textbf{\textcolor{citecolor}{+0.098}} 
& 29.432 & \textbf{\textcolor{citecolor}{+0.055}} 
& 29.546 & \textbf{\textcolor{citecolor}{+0.309}} 
& 35.109 & \textbf{\textcolor{citecolor}{+0.226}}
& 31.974 & \textbf{\textcolor{citecolor}{+0.149}} \\
{\bf +} Larger Patches 
& {\bf 34.940} & \textbf{\textcolor{citecolor}{+0.005}}
& {\bf 30.858} & \textbf{\textcolor{citecolor}{+0.011}}
& {\bf 29.439} & \textbf{\textcolor{citecolor}{+0.007}}
& {\bf 29.567} & \textbf{\textcolor{citecolor}{+0.021}}	
& {\bf 35.132} & \textbf{\textcolor{citecolor}{+0.023}}
& {\bf 31.987} & \textbf{\textcolor{citecolor}{+0.013}}\\
\midrule
Warm Start ($\times\textbf{4}$ SR) 
& 32.719 & - & 28.986 & - & 27.847 & - & 27.024 & - & 31.676 & - & 29.650 & - \\
{\bf +} Normal Finetuning
& 32.805 & \textbf{\textcolor{citecolor}{+0.086}}
& 29.077 & \textbf{\textcolor{citecolor}{+0.091}} 
& 27.909 & \textbf{\textcolor{citecolor}{+0.062}} 
& 27.325 & \textbf{\textcolor{citecolor}{+0.301}} 
& 32.014 & \textbf{\textcolor{citecolor}{+0.338}}
& 29.826 & \textbf{\textcolor{citecolor}{+0.176}} \\
{\bf +} Larger Patches 
& {\bf 32.809} & \textbf{\textcolor{citecolor}{+0.004}}
& {\bf 29.081} & \textbf{\textcolor{citecolor}{+0.004}}
& {\bf 27.912} & \textbf{\textcolor{citecolor}{+0.003}}
& {\bf 27.338} & \textbf{\textcolor{citecolor}{+0.013}}	
& {\bf 32.035} & \textbf{\textcolor{citecolor}{+0.021}}
& {\bf 29.835} & \textbf{\textcolor{citecolor}{+0.009}}\\
\bottomrule
\end{tabular}
} 
\end{center}
\end{table*}

\begin{figure*}[t]
\scriptsize
\centering
\begin{tabular}{cc}
\hspace{-0.4cm}
\begin{adjustbox}{valign=t}
\begin{tabular}{c}
\includegraphics[width=0.197\textwidth]{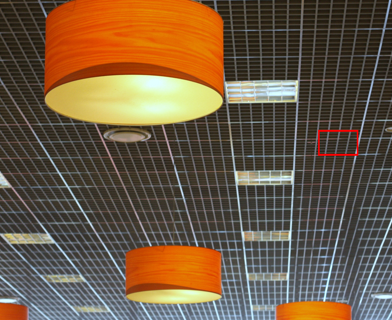}
\\
Urban100: img\_044 ($\times$4)
\end{tabular}
\end{adjustbox}
\hspace{-0.46cm}
\begin{adjustbox}{valign=t}
\begin{tabular}{cccccc}
\includegraphics[width=0.138\textwidth]{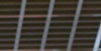} \hspace{-4mm} &
\includegraphics[width=0.138\textwidth]{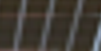} \hspace{-4mm} &
\includegraphics[width=0.138\textwidth]{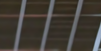} \hspace{-4mm} &
\includegraphics[width=0.138\textwidth]{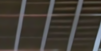} \hspace{-4mm} &
\includegraphics[width=0.138\textwidth]{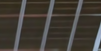} \hspace{-4mm} 
\\
HQ \hspace{-4mm} &
Bicubic \hspace{-4mm} &
EDSR~\cite{lim2017enhanced} \hspace{-4mm} &
RDN~\cite{zhang2018residual} \hspace{-4mm} &
RCAN~\cite{zhang2018image} \hspace{-4mm}
\\
\includegraphics[width=0.138\textwidth]{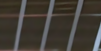} \hspace{-4mm} &
\includegraphics[width=0.138\textwidth]{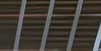} \hspace{-4mm} &
\includegraphics[width=0.138\textwidth]{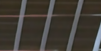} \hspace{-4mm} &
\includegraphics[width=0.138\textwidth]{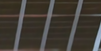} \hspace{-4mm} &
\includegraphics[width=0.138\textwidth]{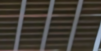} \hspace{-4mm}  
\\ 
SRFBN~\cite{li2019feedback} \hspace{-4mm} &
IGNN~\cite{zhou2020cross} \hspace{-4mm} &
CSNLN~\cite{mei2020image} \hspace{-4mm} &
RFANet~\cite{liu2020residual}  \hspace{-4mm} &
RCAN-it (ours) \hspace{-4mm}
\\
\end{tabular}
\end{adjustbox}
\vspace{1mm}
\\
\hspace{-0.4cm}
\begin{adjustbox}{valign=t}
\begin{tabular}{c}
\includegraphics[width=0.197\textwidth]{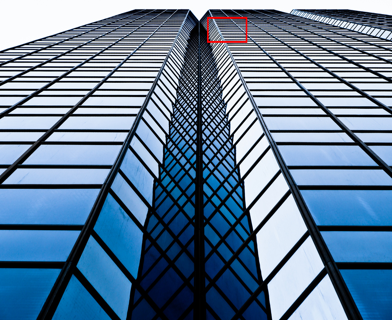}
\\
Urban100: img\_067 ($\times$4)
\end{tabular}
\end{adjustbox}
\hspace{-0.46cm}
\begin{adjustbox}{valign=t}
\begin{tabular}{cccccc}
\includegraphics[width=0.138\textwidth]{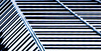} \hspace{-4mm} &
\includegraphics[width=0.138\textwidth]{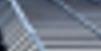} \hspace{-4mm} &
\includegraphics[width=0.138\textwidth]{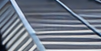} \hspace{-4mm} &
\includegraphics[width=0.138\textwidth]{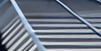} \hspace{-4mm} &
\includegraphics[width=0.138\textwidth]{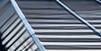} \hspace{-4mm} 
\\
HQ \hspace{-4mm} &
Bicubic \hspace{-4mm} &
EDSR~\cite{lim2017enhanced} \hspace{-4mm} &
RDN~\cite{zhang2018residual} \hspace{-4mm} &
RCAN~\cite{zhang2018image} \hspace{-4mm}
\\
\includegraphics[width=0.138\textwidth]{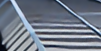} \hspace{-4mm} &
\includegraphics[width=0.138\textwidth]{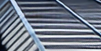} \hspace{-4mm} &
\includegraphics[width=0.138\textwidth]{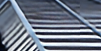} \hspace{-4mm} &
\includegraphics[width=0.138\textwidth]{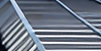} \hspace{-4mm} &
\includegraphics[width=0.138\textwidth]{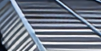} \hspace{-4mm}  
\\ 
SRFBN~\cite{li2019feedback} \hspace{-4mm} &
IGNN~\cite{zhou2020cross} \hspace{-4mm} &
CSNLN~\cite{mei2020image} \hspace{-4mm} &
RFANet~\cite{liu2020residual}  \hspace{-4mm} &
RCAN-it (ours) \hspace{-4mm}
\\
\end{tabular}
\end{adjustbox}
\vspace{1mm}
\\
\hspace{-0.4cm}
\begin{adjustbox}{valign=t}
\begin{tabular}{c}
\includegraphics[width=0.197\textwidth]{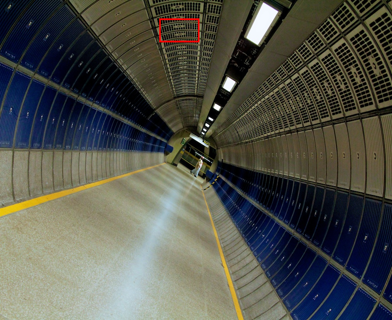}
\\
Urban100: img\_078 ($\times$4)
\end{tabular}
\end{adjustbox}
\hspace{-0.46cm}
\begin{adjustbox}{valign=t}
\begin{tabular}{cccccc}
\includegraphics[width=0.138\textwidth]{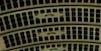} \hspace{-4mm} &
\includegraphics[width=0.138\textwidth]{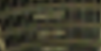} \hspace{-4mm} &
\includegraphics[width=0.138\textwidth]{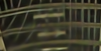} \hspace{-4mm} &
\includegraphics[width=0.138\textwidth]{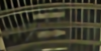} \hspace{-4mm} &
\includegraphics[width=0.138\textwidth]{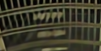} \hspace{-4mm} 
\\
HQ \hspace{-4mm} &
Bicubic \hspace{-4mm} &
EDSR~\cite{lim2017enhanced} \hspace{-4mm} &
RDN~\cite{zhang2018residual} \hspace{-4mm} &
RCAN~\cite{zhang2018image} \hspace{-4mm}
\\
\includegraphics[width=0.138\textwidth]{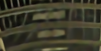} \hspace{-4mm} &
\includegraphics[width=0.138\textwidth]{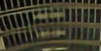} \hspace{-4mm} &
\includegraphics[width=0.138\textwidth]{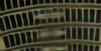} \hspace{-4mm} &
\includegraphics[width=0.138\textwidth]{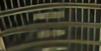} \hspace{-4mm} &
\includegraphics[width=0.138\textwidth]{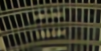} \hspace{-4mm}  
\\ 
SRFBN~\cite{li2019feedback} \hspace{-4mm} &
IGNN~\cite{zhou2020cross} \hspace{-4mm} &
CSNLN~\cite{mei2020image} \hspace{-4mm} &
RFANet~\cite{liu2020residual}  \hspace{-4mm} &
RCAN-it (ours) \hspace{-4mm}
\\
\end{tabular}
\end{adjustbox}
\vspace{1mm}
\\
\hspace{-0.45cm}
\begin{adjustbox}{valign=t}
\begin{tabular}{c}
\includegraphics[width=0.197\textwidth]{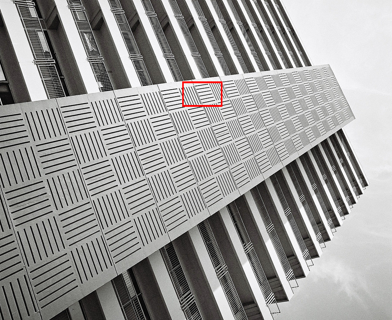}
\\
Urban100: img\_092 ($\times$4)
\end{tabular}
\end{adjustbox}
\hspace{-0.46cm}
\begin{adjustbox}{valign=t}
\begin{tabular}{cccccc}
\includegraphics[width=0.138\textwidth]{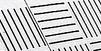} \hspace{-4mm} &
\includegraphics[width=0.138\textwidth]{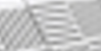} \hspace{-4mm} &
\includegraphics[width=0.138\textwidth]{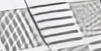} \hspace{-4mm} &
\includegraphics[width=0.138\textwidth]{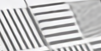} \hspace{-4mm} &
\includegraphics[width=0.138\textwidth]{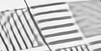} \hspace{-4mm} 
\\
HQ \hspace{-4mm} &
Bicubic \hspace{-4mm} &
EDSR~\cite{lim2017enhanced} \hspace{-4mm} &
RDN~\cite{zhang2018residual} \hspace{-4mm} &
RCAN~\cite{zhang2018image} \hspace{-4mm}
\\
\includegraphics[width=0.138\textwidth]{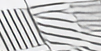} \hspace{-4mm} &
\includegraphics[width=0.138\textwidth]{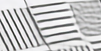} \hspace{-4mm} &
\includegraphics[width=0.138\textwidth]{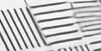} \hspace{-4mm} &
\includegraphics[width=0.138\textwidth]{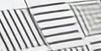} \hspace{-4mm} &
\includegraphics[width=0.138\textwidth]{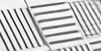} \hspace{-4mm}  
\\ 
SRFBN~\cite{li2019feedback} \hspace{-4mm} &
IGNN~\cite{zhou2020cross} \hspace{-4mm} &
CSNLN~\cite{mei2020image} \hspace{-4mm} &
RFANet~\cite{liu2020residual}  \hspace{-4mm} &
RCAN-it (ours) \hspace{-4mm}
\\
\end{tabular}
\end{adjustbox}

\end{tabular}
\caption{Visual comparison ($\times$4) with large SR networks on the Urban100~\cite{huang2015single} dataset. Our simply improved RCAN-it can better reconstruct high-frequency details than RCAN~\cite{zhang2018image} and also compare favorably with previous state-of-the-art approaches. 
}
\label{fig:rcanit_visual_result_SRBIX4_large}
\end{figure*}

\begin{table*}[thbp]
\footnotesize
\center
\begin{center}
\caption{
Quantitative comparison with CNN-based SR approaches. We show the PSNR (dB) and SSIM for $\times2$, $\times3$ and $\times4$ SR tasks on standard benchmark datasets. Our simply improved RCAN-{\em it} outperforms or matches existing CNN-based models on almost all metrics, demonstrating the practical value of the improved training strategies. Symbol {\bf+} indicates results produced with self-ensemble at inference time (note that DFSA~\cite{magid2021dynamic} only reports self-ensemble scores).
}
\label{tab:all_results_psnr_ssim_x2x3x4}
\resizebox{0.98\textwidth}{!}{%
\begin{tabular}{|l|c|c|c|c|c|c|c|c|c|c|c|c|}
\hline
\multirow{2}{*}{Method} & \multirow{2}{*}{Year} & \multirow{2}{*}{Scale} &  \multicolumn{2}{c|}{Set5} &  \multicolumn{2}{c|}{Set14} &  \multicolumn{2}{c|}{B100} &  \multicolumn{2}{c|}{Urban100} &  \multicolumn{2}{c|}{Manga109}  
\\
\cline{4-13}
&  &  & PSNR & SSIM & PSNR & SSIM & PSNR & SSIM & PSNR & SSIM & PSNR & SSIM 
\\
\hline
\hline
                   
EDSR~\cite{lim2017enhanced} & 2017 & $\times$2 
& 38.11
 & 0.9602
  & 33.92
   & 0.9195
    & 32.32
     & 0.9013
      & 32.93
       & 0.9351
        & 39.10
         & 0.9773
                   
\\
DBPN~\cite{haris2018deep} & 2018 & $\times$2 
& 38.09
 & 0.9600
  & 33.85
   & 0.9190
    & 32.27
     & 0.9000
      & 32.55
       & 0.9324
        & 38.89
         & 0.9775        
\\
RDN~\cite{zhang2018residual} & 2018 & $\times$2 
& 38.24
 & 0.9614
  & 34.01
   & 0.9212
    & 32.34
     & 0.9017
      & 32.89
       & 0.9353
        & 39.18
         & 0.9780
         
\\
RCAN~\cite{zhang2018image} & 2018 & $\times$2 
& {38.27}
 & {0.9614}
  & {34.12}
   & {0.9216}
    & {32.41}
     & {0.9027}
      & {33.34}
       & {0.9384}
        & {39.44}
         & {0.9786}
\\       
NLRN~\cite{liu2018non} & 2018 & $\times$2 
& {38.00}
 & {0.9603}
  & {33.46}
   & {0.9159}
    & {32.19}
     & {0.8992}
      & {31.81}
       & {0.9249}
        & -
         & -
\\
RNAN~\cite{zhang2019rnan} & 2019 & $\times$2 
& 38.17
 & 0.9611
  & 33.87
   & 0.9207
    & 32.31
     & 0.9014
      & 32.73
       & 0.9340
        & 39.23
         & 0.9785
         
\\
SRFBN~\cite{li2019feedback} & 2019 & $\times$2 
& {38.11}
 & {0.9609}
  & {33.82}
   & {0.9196}
    & {32.29}
     & {0.9010}
      & {32.62}
       & {0.9328}
        & {39.08}
         &{0.9779}
\\
OISR~\cite{he2019ode} & 2019 & $\times$2 
& {38.21}
 & {0.9612}
  & {33.94}
   & {0.9206}
    & {32.36}
     & {0.9019}
      & {33.03}
       & {0.9365}
        & -
         & -      
\\
SAN~\cite{dai2019second} & 2019 & $\times$2 
& 38.31
 & 0.9620
  & {34.07}
   & {0.9213}
    & 32.42
     & 0.9028
      & {33.10}
       & {0.9370}
        & {39.32}
         & 0.9792
\\
CSNLN~\cite{mei2020image} & 2020 & $\times$2 
& {38.28}
 & 0.9616
  & {34.12}
   & 0.9223
    & {32.40}
     & {0.9024}
      & {33.25}
       & {0.9386}
        & {39.37}
         & {0.9785}
\\
RFANet~\cite{liu2020residual} & 2020 & $\times$2 
& 38.26 
 & 0.9615 
  & 34.16 
   & 0.9220 
    & 32.41 
     & 0.9026 
      & 33.33 
       & 0.9389
        & 39.44 
         & 0.9783
\\
HAN~\cite{niu2020single} & 2020 & $\times$2 
& 38.27 
 & 0.9614 
  & 34.16
   & 0.9217 
    & 32.41 
     & 0.9027 
      & 33.35 
       & 0.9385 
        & 39.46 
         & 0.9785
\\
NSR~\cite{fan2020neural} & 2020 & $\times$2 
& 38.23 
 & 0.9614 
  & 33.94 
   & 0.9203 
    & 32.34 
     & 0.9020 
      & 33.02 
       & 0.9367 
        & 39.31 
         & 0.9782
\\
IGNN~\cite{zhou2020cross} & 2020 & $\times$2 
& 38.24 
 & 0.9613 
  & 34.07 
   & 0.9217 
    & 32.41 
     & 0.9025 
      & 33.23 
       & 0.9383 
        & 39.35 
         & 0.9786
\\
CRAN~\cite{Zhang2021ContextRA} & 2021 & $\times$2 
& 38.31
 & 0.9617
  & 34.22
   & 0.9232
    & 32.44
     & 0.9029
      & 33.43
       & 0.9394
        & 39.75
         & 0.9793
         
\\
DFSA{\bf+}~\cite{magid2021dynamic} & 2021 & $\times$2 
 
& 38.38
    &  0.9620
        &  34.33 
            & 0.9232 
                & 32.50 
                    & 0.9036
                        & 33.66
                            & 0.9412
                                &  39.98 
                                    & 0.9798    
\\  
\hline
RCAN-it  & \multirow{2}{*}{2021} & \multirow{2}{*}{$\times$2}
& {38.37} & {0.9620}
& {34.49} & {0.9250}
& {32.48} & {0.9034}	 
& {33.62} & {0.9410}
& {39.88} & {0.9799}
\\
RCAN-it{\bf+}  & &
& {\bf 38.41} & {\bf 0.9621}
& {\bf 34.55} & {\bf 0.9254}
& {\bf 32.51} & {\bf 0.9038}
& {\bf 33.79} & {\bf 0.9421}
& {\bf 40.04} & {\bf 0.9801} 
\\
\hline
\hline

EDSR~\cite{lim2017enhanced} & 2017 & $\times$3 
& 34.65
 & 0.9280
  & 30.52
   & 0.8462
    & 29.25
     & 0.8093
      & 28.80
       & 0.8653
        & 34.17
         & 0.9476
                   
\\
RDN~\cite{zhang2018residual} & 2018 & $\times$3 
& 34.71
 & 0.9296
  & 30.57
   & 0.8468
    & 29.26
     & 0.8093
      & 28.80
       & 0.8653
        & 34.13
         & 0.9484
         
\\
RCAN~\cite{zhang2018image}& 2018 & $\times$3 
& {34.74}
 &{0.9299}
  & {30.65}
   & {0.8482}
    & {29.32}
     & {0.8111}
      & {29.09}
       &{0.8702}
        & {34.44}
         &{0.9499}
         
\\
NLRN~\cite{liu2018non}& 2018 & $\times$3 
& {34.27}
 &{0.9266}
  & {30.16}
   &{0.8374}
    & {29.06}
     & {0.8026}
      & {27.93}
       & {0.8453}
        & {-}
         & {-}
\\
RNAN~\cite{zhang2019rnan}& 2019 & $\times$3 
& 34.66
 & 0.9290
  & 30.53
   & 0.8463
    & 29.26
     & 0.8090
      & 28.75
       & 0.8646
        & 34.25
         & 0.9483
         
\\
SRFBN~\cite{li2019feedback}& 2019 & $\times$3 
& {34.70}
 &{0.9292}
  & {30.51}
   &{0.8461}
    & {29.24}
     & {0.8084}
      & {28.73}
       & {0.8641}
        & {34.18}
         & {0.9481}
\\
OISR~\cite{he2019ode}& 2019 & $\times$3 
& {34.72}
 &{0.9297}
  & {30.57}
   &{0.8470}
    & {29.29}
     & {0.8103}
      & {28.95}
       & {0.8680}
        & {-}
         & {-}
\\
SAN~\cite{dai2019second} & 2019 & $\times$3 
& {34.75}
 &{0.9300}
  & {30.59}
   &{0.8476}
    &{29.33}
     & {0.8112}
      & {28.93}
       & {0.8671}
        & {34.30}
         & {0.9494}
        
\\
CSNLN~\cite{mei2020image} & 2020 & $\times$3
& {34.74}
 & 0.9300
  & {30.66}
   & {0.8482}
    & {29.33}
     & {0.8105}
      & {29.13}
       & {0.8712}
        & {34.45}
         & {0.9502}
\\
RFANet~\cite{liu2020residual} & 2020 & $\times$3 
& {34.79}
 & {0.9300}
  & {30.67}
   & {0.8487}
    & {29.34}
     & {0.8115}
      & {29.15}
       & {0.8720}
        & {34.59}
         & {0.9506}
         
\\
HAN~\cite{niu2020single} & 2020 & $\times$3 
& 34.75
 & 0.9299
  & 30.67
   & 0.8483
    & 29.32
     & 0.8110
      & 29.10
       & 0.8705
        & 34.48
         & 0.9500
         
\\
NSR~\cite{fan2020neural} & 2020 & $\times$3 
& 34.62
 & 0.9289
  & 30.57
   & 0.8475
    & 29.26
     & 0.8100
      & 28.83
       & 0.8663
        & 34.27
         & 0.9484
         
\\
IGNN~\cite{zhou2020cross} & 2020 & $\times$3 
& 34.72
 & 0.9298
  & 30.66
   & 0.8484
    & 29.31
     & 0.8105
      & 29.03
       & 0.8696
        & 34.39
         & 0.9496
         
\\
CRAN~\cite{Zhang2021ContextRA} & 2021 & $\times$3 
& 34.80
 & 0.9304
  & 30.73
   & 0.8498
    & 29.38
     & 0.8124
      & 29.33
       & 0.8745
        & 34.84
         & 0.9515
         
\\
DFSA{\bf+}~\cite{magid2021dynamic}  & 2021 & $\times$3 
&{34.92} 
    & {0.9312}
        &{30.83}  
            &{0.8507}
                &{29.42}
                    &{0.8128}
                        &{29.44}
                            &{0.8761}
                                &{35.07} 
                                    &{0.9525}    
\\
\hline
RCAN-it  & \multirow{2}{*}{2021} & \multirow{2}{*}{$\times$3} 
& {34.86} & {0.9308}
& {30.76} & {0.8505}
& {29.39} & {0.8125}	 
& {29.38} & {0.8755}
& {34.92} & {0.9520}
\\
RCAN-it{\bf+}  & & 
& {\bf 34.94} & {\bf 0.9313}
& {\bf 30.84} & {\bf 0.8515}
& {\bf 29.44} & {\bf 0.8133}	 
& {\bf 29.57} & {\bf 0.8779}
& {\bf 35.13} & {\bf 0.9528}
\\
\hline
\hline
                   
                   
EDSR~\cite{lim2017enhanced} & 2017 & $\times$4 
& 32.46
 & 0.8968
  & 28.80
   & 0.7876
    & 27.71
     & 0.7420
      & 26.64
       & 0.8033
        & 31.02
         & 0.9148
                   
\\
DBPN~\cite{haris2018deep} & 2018 & $\times$4 
& 32.47
 & 0.8980
  & 28.82
   & 0.7860
    & 27.72
     & 0.7400
      & 26.38
       & 0.7946
        & 30.91
         & 0.9137
         
\\
RDN~\cite{zhang2018residual} & 2018 & $\times$4 
& 32.47
 & 0.8990
  & 28.81
   & 0.7871
    & 27.72
     & 0.7419
      & 26.61
       & 0.8028
        & 31.00
         & 0.9151
         
\\
RCAN~\cite{zhang2018image}& 2018 & $\times$4 
& {32.63}
 & {0.9002}
  & {28.87}
   &{0.7889}
    & {27.77}
     & {0.7436}
      & {26.82}
       & {0.8087}
        &{31.22}
         & {0.9173}

\\
NLRN~\cite{liu2018non}& 2018 & $\times$4 
& {31.92}
 & {0.8916}
  & {28.36}
   & {0.7745}
    & {27.48}
     & {0.7306}
      & {25.79}
       & {0.7729}
        & {-}
         & {-}
\\
RNAN~\cite{zhang2019rnan}& 2019 & $\times$4 
& 32.43
 & 0.8977
  & 28.83
   & 0.7871
    & 27.72
     & 0.7410
      & 26.61
       & 0.8023
        & 31.09
         & 0.9149
         
\\
SRFBN~\cite{li2019feedback} & 2019 & $\times$4 
& {32.47}
 & {0.8983}
  & {28.81}
   & {0.7868}
    & {27.72}
     & {0.7409}
      & {26.60}
       & {0.8015}
        & {31.15}
         & {0.9160}
\\
OISR~\cite{he2019ode} & 2019 & $\times$4 
&{32.53}
 &{0.8992}
  &{28.86}
   & {0.7878}
    &{27.75}
     & {0.7428}
      & {26.79}
       & {0.8068}
        & {-}
         & {-}
\\
SAN~\cite{dai2019second} & 2019 & $\times$4 
& {32.64}
 &{0.9003}
  &{28.92}
   &{0.7888}
    &{27.78}
     & {0.7436}
      & {26.79}
       & {0.8068}
        & {31.18}
         & {0.9169}
\\
CSNLN~\cite{mei2020image}  & 2020 & $\times$4 
& {32.68}
 & {0.9004}
  & {28.95}
   & {0.7888}
    & {27.80}
     & {0.7439}
      & {27.22}
       & {0.8168}
        & {31.43}
         & {0.9201}
\\
RFANet~\cite{liu2020residual} & 2020 & $\times$4 
& 32.66
 & 0.9004
  & 28.88
   & {0.7894}
    & 27.79
     & 0.7442
      & 26.92
       & 0.8112
        & 31.41
         & 0.9187
         
\\
HAN~\cite{niu2020single} & 2020 & $\times$4 
& 32.64
 & 0.9002
  & 28.90
   & 0.7890
    & {27.80}
     & {0.7442}
      & 26.85
       & 0.8094
        & 31.42 
         & 0.9177
         
\\
NSR~\cite{fan2020neural} & 2020 & $\times$4 
& 32.55
 & 0.8987
  & 28.79
   & 0.7876
    & 27.72
     & 0.7414
      & 26.61
       & 0.8025
        & 31.10
         & 0.9145
         
\\
IGNN~\cite{zhou2020cross} & 2020 & $\times$4 
& 32.57
 & 0.8998
  & 28.85
   & 0.7891
    & 27.77
     & 0.7434
      & 26.84
       & 0.8090
        & 31.28
         & 0.9182
         
\\
CRAN~\cite{Zhang2021ContextRA} & 2021 & $\times$4 
& 32.72
 & 0.9012
  & 29.01
   & 0.7918
    & 27.86
     & 0.7460
      & 27.13
       & 0.8167
        & 31.75
         & 0.9219
         
\\
DFSA{\bf+}~\cite{magid2021dynamic} & 2021 & $\times$4
& {32.79}
    &{0.9019} 
        & {29.06}
            & {0.7922 }
                & {27.87}
                    & {0.7458}
                        &{27.17}
                            & {0.8163} 
                                & {31.88}
                                    & {\bf 0.9266}   \\
\hline
RCAN-it & \multirow{2}{*}{2021} & \multirow{2}{*}{$\times$4} 
& {32.69} & {0.9007}
& {28.99} & {0.7922}
& {27.87} & {0.7459}	 
& {27.16} & {0.8168}
& {31.78} & {0.9217}
\\
RCAN-it{\bf+} & &
& {\bf 32.81} & {\bf 0.9021}
& {\bf 29.08} & {\bf 0.7933}
& {\bf 27.91} & {\bf 0.7469}
& {\bf 27.34} & {\bf 0.8201}
& {\bf 32.04} & 0.9233
\\                                    
\hline             
\end{tabular}
} 
\end{center}
\label{tab:all_comparisons}
\end{table*}

\begin{table*}[t]
\center
\begin{center}
\caption{
Quantitative comparison with SwinIR~\cite{liang2021swinir}, a transformer-based SR model. PSNR (dB) and SSIM are shown for the $\times2$ SR task on the Manga109 dataset. Inference time are benchmarked on a Nvidia V100 GPU for $\times2$ SR with $256\times256$ input patches. SwinIR-S denotes a lightweight version of the standard SwinIR model. Symbol {\bf+} denotes self-ensemble results.
}
\label{tab:swinIR}
\resizebox{\textwidth}{!}{%
\begin{tabular}{lcccccccccccc}
\toprule
\multirow{2}{*}{Methods}
& \multirow{2}{*}{Time (Sec.)}
& \multirow{2}{*}{Rel. RCAN}
& \multicolumn{2}{c}{Set5} 
& \multicolumn{2}{c}{Set14} 
& \multicolumn{2}{c}{B100} 
& \multicolumn{2}{c}{Urban100} 
& \multicolumn{2}{c}{Manga109}
\\
\cmidrule{4-13}
& &
& PSNR & SSIM
& PSNR & SSIM
& PSNR & SSIM
& PSNR & SSIM
& PSNR & SSIM
\\
\midrule
RCAN{\bf+}~\cite{zhang2018image} & 0.1645 & 1.00 & 38.33 & 0.9617 & 34.23 & 0.9225 & 32.46 & 0.9031 & 33.54 & 0.9399 & 39.61 & 0.9788\\
SwinIR{\bf+}~\cite{liang2021swinir} & 0.7137 & 4.34 & 38.46 & 0.9624 & 34.61 & 0.9260 & 32.55 & 0.9043 & 33.95 & 0.9433 & 40.02 & 0.9800\\
SwinIR-S{\bf+}~\cite{liang2021swinir} & 0.4747 & 2.89 & 38.19 & 0.9613 & 33.96 & 0.9211 & 32.34 & 0.9015 & 32.90 & 0.9349 & 39.23 & 0.9783\\
\midrule
RCAN-it{\bf+} ({\bf Ours}) & 0.1655 & 1.01 &  38.41 & 0.9621
& 34.55 & 0.9254
& 32.51 & 0.9038
& 33.79 & 0.9421
& 40.04 & 0.9801 \\
\bottomrule
\end{tabular}
} 
\end{center}
\end{table*}

\bfsection{Warm start} As described before, we start from a pre-trained $\times2$ SR model to initialize the weights of the $\times3$ and $\times4$ models as they share the training data (just with different down-sampling ratios), which can save training cost. By finetuning only the \texttt{tail} module while keeping other layers frozen, $\times3$ and $\times4$ models can already match or be slightly lower than the original results in the RCAN paper (Table~\ref{tab:warm_start}). This shows that warm-start is an effective and efficient way to construct models with different SR scales. Finetuning the whole model for 80K iterations ($50\%$ of the longer training setting in Table~\ref{tab:increment}) can boost the average performance by 0.149 dB for $\times3$ SR and 0.176 dB for $\times 4$ SR. We also notice that large-patch finetuning still increases the SR scores for both scales (Table~\ref{tab:warm_start}), but the improvement is relatively marginal compared with the $\times2$ case. Our results suggest that using warm start with finetuning instead of training from scratch can be a reasonable practice for real-world applications with a cost (time) budget.

\subsection{Benchmark Comparison}\label{sec:benchmark}

So far, we have identified a proper training protocol to improve RCAN, which has been released for more than three years. In this section, we compare our RCAN-it with existing CNN-based and transformer-based SR models that have defined the state-of-the-art in the past several years and discuss the pros and cons of existing approaches. 

\bfsection{Visual comparison} In a visual comparison (Fig.~\ref{fig:rcanit_visual_result_SRBIX4_large}) of $\times4$ SR on Urban100, we show that RCAN-it can reconstruct high-frequency details better than not only RCAN~\cite{zhang2018image}, but also more recent works including IGNN~\cite{zhou2020cross}, CSNLN~\cite{mei2020image} and RFANet~\cite{liu2020residual}. Specifically, RCAN-it reconstructs the strip patterns with higher contrast, while other approaches tend to hallucinate unrealistic artifacts (Fig.~\ref{fig:rcanit_visual_result_SRBIX4_large}, 4th row). Besides, structures with lower contrast from the background can be eliminated in previous models, while our RCAN-it can preserve them (Fig.~\ref{fig:rcanit_visual_result_SRBIX4_large}, 1st row). Please note that except for the small modification of activation function, our RCAN-it compares favorably with existing approaches in reconstructing high-frequency details just with better {\em training} instead of introducing new modules or structures exclusively for learning high-frequency components.

\bfsection{Quantitative comparison} Table~\ref{tab:all_results_psnr_ssim_x2x3x4} shows the comprehensive quantitative comparison of RCAN-it with existing CNN-based SR approaches released in the past four years. The numbers are without self-ensemble by default. DFSA~\cite{magid2021dynamic} only reports self-ensemble scores, therefore a fair comparison is DFSA{\bf+} \vs RCAN-it{\bf+}. The quantitative results show that even for models released three years after RCAN, our simply improved RCAN-it can outperform or match the performance, with or without self-ensemble. Please note that we are {\em not} claiming the advance in SR architectures is not important. On the contrary, we champion the success of novel architectures that pushes the field forward. Our results show that similar or higher empirical improvement can be achieved by upgrading the training, and we believe the practical optimization solution we provided can help the latest models shine in the field.

\bfsection{Comparison with SwinIR} Finally, we show the comparison with a concurrent work, SwinIR~\cite{liang2021swinir}, which adapts Swin Transformer~\cite{liu2021swin} for image restoration (Table~\ref{tab:swinIR}). SwinIR is the current state-of-the-art to the best of our knowledge, outperforming RCAN by 0.41 dB on both Urban100 and Manga109 for the $\times2$ SR task. With an improved training strategy, our RCAN-it has clearly narrowed the gap between RCAN and SwinIR on Urban100 and even outperformed SwinIR on Manga109. Besides, we show that despite the impressive SR performance, SwinIR suffers from a slow inference speed. When benchmarking on a single Nvidia V100 GPU, the standard version of SwinIR is $4.34\times$ slower than RCAN, and even a lightweight version SwinIR-S is still $2.89\times$ slower but with significantly decreased performance. Compared with SwinIR, the $1\%$ extra inference time due to SiLU in our RCAN-it is negligible. In real-world applications where SR quality and inference speed are weighted differently, RCAN-it will still be a reasonable choice considering the quality-speed tradeoff.

\subsection{Discussion} 
We have compared RCAN-it with existing SR models and demonstrate the simplicity but surprising effectiveness of updated training strategies. The oracle study also indicates the large space for improving SR models from different aspects besides architecture (Table.~\ref{tab:increment}). One current limitation is that we have not tested our improved training protocol with the latest models. We believe at least recent CNN-based models can enjoy performance gain by adapting our solution and leave that for future exploration. We also want to emphasize that the {\em underfitting} problem and the empirical improvements are strongly correlated with the DF2K~\cite{lim2017enhanced} dataset for training the models. Researchers may have different observations using different training data.

\section{Conclusion and Future Work} 
In this work, we revisit the standard RCAN~\cite{zhang2018image} and demonstrate that with a proper training strategy, we can alleviate the {\em underfitting} problem and increase the image super-resolution performance by a large margin. This update makes our RCAN-it (RCAN with improved training) better than or comparable with the latest CNN-based approaches, demonstrating the essentiality of training besides architecture. Recently, the popularity of vision transformer~\cite{dosovitskiy2020image} has inspired a wave of transformer-based architectures in both high-level~\cite{dosovitskiy2020image,liu2021swin} and low-level~\cite{liang2021swinir} tasks. However, Dosovitskiy \etal~\cite{dosovitskiy2020image} shows that the optimal setting for training CNN-based classifiers and vision transformers can be very different. In future research, we will also explore the training of transformers in SR.

\bfsection{Acknowledgements} We thank the support from NSF award IIS-2124179 and NIH award 5U54CA225088-03.

{\small
\bibliographystyle{ieee_fullname}
\bibliography{egbib}

\begin{thebibliography}{10}\itemsep=-1pt

\bibitem{ash2020warm}
Jordan Ash and Ryan~P Adams.
\newblock On warm-starting neural network training.
\newblock {\em Advances in Neural Information Processing Systems}, 33, 2020.

\bibitem{bello2021revisiting}
Irwan Bello, William Fedus, Xianzhi Du, Ekin~D Cubuk, Aravind Srinivas,
  Tsung-Yi Lin, Jonathon Shlens, and Barret Zoph.
\newblock Revisiting resnets: Improved training and scaling strategies.
\newblock {\em arXiv preprint arXiv:2103.07579}, 2021.

\bibitem{bevilacqua2012low}
Marco Bevilacqua, Aline Roumy, Christine Guillemot, and Marie~Line
  Alberi-Morel.
\newblock Low-complexity single-image super-resolution based on nonnegative
  neighbor embedding.
\newblock In {\em BMVC}, 2012.

\bibitem{cubuk2020randaugment}
Ekin~D Cubuk, Barret Zoph, Jonathon Shlens, and Quoc~V Le.
\newblock Randaugment: Practical automated data augmentation with a reduced
  search space.
\newblock In {\em CVPRW}, 2020.

\bibitem{dai2017deformable}
Jifeng Dai, Haozhi Qi, Yuwen Xiong, Yi Li, Guodong Zhang, Han Hu, and Yichen
  Wei.
\newblock Deformable convolutional networks.
\newblock In {\em ICCV}, 2017.

\bibitem{dai2019second}
Tao Dai, Jianrui Cai, Yongbing Zhang, Shu-Tao Xia, and Lei Zhang.
\newblock Second-order attention network for single image super-resolution.
\newblock In {\em CVPR}, 2019.

\bibitem{dai2019imageRDAN}
Tao Dai, Hua Zha, Yong Jiang, and Shu-Tao Xia.
\newblock Image super-resolution via residual block attention networks.
\newblock In {\em CVPRW}, 2019.

\bibitem{dong2016image}
Chao Dong, Chen~Change Loy, Kaiming He, and Xiaoou Tang.
\newblock Image super-resolution using deep convolutional networks.
\newblock {\em TPAMI}, 2016.

\bibitem{dong2016accelerating}
Chao Dong, Chen~Change Loy, and Xiaoou Tang.
\newblock Accelerating the super-resolution convolutional neural network.
\newblock In {\em ECCV}, 2016.

\bibitem{dosovitskiy2020image}
Alexey Dosovitskiy, Lucas Beyer, Alexander Kolesnikov, Dirk Weissenborn,
  Xiaohua Zhai, Thomas Unterthiner, Mostafa Dehghani, Matthias Minderer, Georg
  Heigold, Sylvain Gelly, et~al.
\newblock An image is worth 16x16 words: Transformers for image recognition at
  scale.
\newblock {\em ICLR}, 2021.

\bibitem{fan2020neural}
Yuchen Fan, Jiahui Yu, Yiqun Mei, Yulun Zhang, Yun Fu, Ding Liu, and Thomas~S
  Huang.
\newblock Neural sparse representation for image restoration.
\newblock In {\em NeurIPS}, 2020.

\bibitem{freeman2002example}
William~T Freeman, Thouis~R Jones, and Egon~C Pasztor.
\newblock Example-based super-resolution.
\newblock {\em IEEE Computer graphics and Applications}, 22(2):56--65, 2002.

\bibitem{goyal2017accurate}
Priya Goyal, Piotr Doll{\'a}r, Ross Girshick, Pieter Noordhuis, Lukasz
  Wesolowski, Aapo Kyrola, Andrew Tulloch, Yangqing Jia, and Kaiming He.
\newblock Accurate, large minibatch sgd: Training imagenet in 1 hour.
\newblock {\em arXiv preprint arXiv:1706.02677}, 2017.

\bibitem{haris2018deep}
Muhammad Haris, Greg Shakhnarovich, and Norimichi Ukita.
\newblock Deep back-projection networks for super-resolution.
\newblock In {\em CVPR}, 2018.

\bibitem{he2016deep}
Kaiming He, Xiangyu Zhang, Shaoqing Ren, and Jian Sun.
\newblock Deep residual learning for image recognition.
\newblock In {\em CVPR}, 2016.

\bibitem{he2019bag}
Tong He, Zhi Zhang, Hang Zhang, Zhongyue Zhang, Junyuan Xie, and Mu Li.
\newblock Bag of tricks for image classification with convolutional neural
  networks.
\newblock In {\em CVPR}, 2019.

\bibitem{he2019ode}
Xiangyu He, Zitao Mo, Peisong Wang, Yang Liu, Mingyuan Yang, and Jian Cheng.
\newblock Ode-inspired network design for single image super-resolution.
\newblock In {\em CVPR}, 2019.

\bibitem{hendrycks2016gaussian}
Dan Hendrycks and Kevin Gimpel.
\newblock Gaussian error linear units (gelus).
\newblock {\em arXiv preprint arXiv:1606.08415}, 2016.

\bibitem{hu2018squeeze}
Jie Hu, Li Shen, and Gang Sun.
\newblock Squeeze-and-excitation networks.
\newblock In {\em CVPR}, 2018.

\bibitem{huang2016deep}
Gao Huang, Yu Sun, Zhuang Liu, Daniel Sedra, and Kilian~Q Weinberger.
\newblock Deep networks with stochastic depth.
\newblock In {\em ECCV}, 2016.

\bibitem{huang2015single}
Jia-Bin Huang, Abhishek Singh, and Narendra Ahuja.
\newblock Single image super-resolution from transformed self-exemplars.
\newblock In {\em CVPR}, 2015.

\bibitem{kim2016accurate}
Jiwon Kim, Jung Kwon~Lee, and Kyoung Mu~Lee.
\newblock Accurate image super-resolution using very deep convolutional
  networks.
\newblock In {\em CVPR}, 2016.

\bibitem{kingma2014adam}
Diederik Kingma and Jimmy Ba.
\newblock Adam: A method for stochastic optimization.
\newblock In {\em ICLR}, 2014.

\bibitem{ledig2017photo}
Christian Ledig, Lucas Theis, Ferenc Husz{\'a}r, Jose Caballero, Andrew
  Cunningham, Alejandro Acosta, Andrew Aitken, Alykhan Tejani, Johannes Totz,
  Zehan Wang, and Wenzhe Shi.
\newblock Photo-realistic single image super-resolution using a generative
  adversarial network.
\newblock In {\em CVPR}, 2017.

\bibitem{li2019feedback}
Zhen Li, Jinglei Yang, Zheng Liu, Xiaomin Yang, Gwanggil Jeon, and Wei Wu.
\newblock Feedback network for image super-resolution.
\newblock In {\em CVPR}, 2019.

\bibitem{liang2021swinir}
Jingyun Liang, Jiezhang Cao, Guolei Sun, Kai Zhang, Luc Van~Gool, and Radu
  Timofte.
\newblock Swinir: Image restoration using swin transformer.
\newblock In {\em ICCVW}, 2021.

\bibitem{lim2017enhanced}
Bee Lim, Sanghyun Son, Heewon Kim, Seungjun Nah, and Kyoung~Mu Lee.
\newblock Enhanced deep residual networks for single image super-resolution.
\newblock In {\em CVPRW}, 2017.

\bibitem{lipton2018troubling}
Zachary~C Lipton and Jacob Steinhardt.
\newblock Troubling trends in machine learning scholarship.
\newblock {\em arXiv preprint arXiv:1807.03341}, 2018.

\bibitem{liu2018non}
Ding Liu, Bihan Wen, Yuchen Fan, Chen~Change Loy, and Thomas~S Huang.
\newblock Non-local recurrent network for image restoration.
\newblock In {\em NeurIPS}, 2018.

\bibitem{liu2020residual}
Jie Liu, Wenjie Zhang, Yuting Tang, Jie Tang, and Gangshan Wu.
\newblock Residual feature aggregation network for image super-resolution.
\newblock In {\em CVPR}, 2020.

\bibitem{liu2021swin}
Ze Liu, Yutong Lin, Yue Cao, Han Hu, Yixuan Wei, Zheng Zhang, Stephen Lin, and
  Baining Guo.
\newblock Swin transformer: Hierarchical vision transformer using shifted
  windows.
\newblock {\em arXiv preprint arXiv:2103.14030}, 2021.

\bibitem{loshchilov2016sgdr}
Ilya Loshchilov and Frank Hutter.
\newblock Sgdr: Stochastic gradient descent with warm restarts.
\newblock {\em arXiv preprint arXiv:1608.03983}, 2016.

\bibitem{magid2021dynamic}
Salma~Abdel Magid, Yulun Zhang, Donglai Wei, Won-Dong Jang, Zudi Lin, Yun Fu,
  and Hanspeter Pfister.
\newblock Dynamic high-pass filtering and multi-spectral attention for image
  super-resolution.
\newblock In {\em ICCV}, 2021.

\bibitem{martin2001database}
David Martin, Charless Fowlkes, Doron Tal, and Jitendra Malik.
\newblock A database of human segmented natural images and its application to
  evaluating segmentation algorithms and measuring ecological statistics.
\newblock In {\em ICCV}, 2001.

\bibitem{matsui2017sketch}
Yusuke Matsui, Kota Ito, Yuji Aramaki, Azuma Fujimoto, Toru Ogawa, Toshihiko
  Yamasaki, and Kiyoharu Aizawa.
\newblock Sketch-based manga retrieval using manga109 dataset.
\newblock {\em Multimedia Tools and Applications}, 2017.

\bibitem{mei2020image}
Yiqun Mei, Yuchen Fan, Yuqian Zhou, Lichao Huang, Thomas~S Huang, and Humphrey
  Shi.
\newblock Image super-resolution with cross-scale non-local attention and
  exhaustive self-exemplars mining.
\newblock In {\em CVPR}, 2020.

\bibitem{niu2020single}
Ben Niu, Weilei Wen, Wenqi Ren, Xiangde Zhang, Lianping Yang, Shuzhen Wang,
  Kaihao Zhang, Xiaochun Cao, and Haifeng Shen.
\newblock Single image super-resolution via a holistic attention network.
\newblock In {\em ECCV}, 2020.

\bibitem{paszke2019pytorch}
Adam Paszke, Sam Gross, Francisco Massa, Adam Lerer, James Bradbury, Gregory
  Chanan, Trevor Killeen, Zeming Lin, Natalia Gimelshein, Luca Antiga, et~al.
\newblock Pytorch: An imperative style, high-performance deep learning library.
\newblock {\em NeurIPS}, 2019.

\bibitem{Ramachandran2017SwishAS}
Prajit Ramachandran, Barret Zoph, and Quoc~V. Le.
\newblock Swish: a self-gated activation function.
\newblock {\em arXiv: Neural and Evolutionary Computing}, 2017.

\bibitem{shi2016real}
Wenzhe Shi, Jose Caballero, Ferenc Husz{\'a}r, Johannes Totz, Andrew~P Aitken,
  Rob Bishop, Daniel Rueckert, and Zehan Wang.
\newblock Real-time single image and video super-resolution using an efficient
  sub-pixel convolutional neural network.
\newblock In {\em CVPR}, 2016.

\bibitem{sun2021autoflow}
Deqing Sun, Daniel Vlasic, Charles Herrmann, Varun Jampani, Michael Krainin,
  Huiwen Chang, Ramin Zabih, William~T Freeman, and Ce Liu.
\newblock Autoflow: Learning a better training set for optical flow.
\newblock In {\em CVPR}, 2021.

\bibitem{sun2019models}
Deqing Sun, Xiaodong Yang, Ming-Yu Liu, and Jan Kautz.
\newblock Models matter, so does training: An empirical study of cnns for
  optical flow estimation.
\newblock {\em TPAMI}, 2019.

\bibitem{timofte2017ntire}
Radu Timofte, Eirikur Agustsson, Luc Van~Gool, Ming-Hsuan Yang, Lei Zhang, Bee
  Lim, Sanghyun Son, Heewon Kim, Seungjun Nah, Kyoung~Mu Lee, et~al.
\newblock Ntire 2017 challenge on single image super-resolution: Methods and
  results.
\newblock In {\em CVPRW}, 2017.

\bibitem{vaswani2017attention}
Ashish Vaswani, Noam Shazeer, Niki Parmar, Jakob Uszkoreit, Llion Jones,
  Aidan~N Gomez, {\L}ukasz Kaiser, and Illia Polosukhin.
\newblock Attention is all you need.
\newblock In {\em NeurIPS}, 2017.

\bibitem{wang2004image}
Zhou Wang, Alan~C Bovik, Hamid~R Sheikh, and Eero~P Simoncelli.
\newblock Image quality assessment: from error visibility to structural
  similarity.
\newblock {\em TIP}, 2004.

\bibitem{you2017large}
Yang You, Igor Gitman, and Boris Ginsburg.
\newblock Large batch training of convolutional networks.
\newblock {\em arXiv preprint arXiv:1708.03888}, 2017.

\bibitem{you2019large}
Yang You, Jing Li, Sashank Reddi, Jonathan Hseu, Sanjiv Kumar, Srinadh
  Bhojanapalli, Xiaodan Song, James Demmel, Kurt Keutzer, and Cho-Jui Hsieh.
\newblock Large batch optimization for deep learning: Training bert in 76
  minutes.
\newblock {\em arXiv preprint arXiv:1904.00962}, 2019.

\bibitem{zeyde2012single}
Roman Zeyde, Michael Elad, and Matan Protter.
\newblock On single image scale-up using sparse-representations.
\newblock In {\em Proc. 7th Int. Conf. Curves Surf.}, 2010.

\bibitem{zhang2017mixup}
Hongyi Zhang, Moustapha Cisse, Yann~N Dauphin, and David Lopez-Paz.
\newblock mixup: Beyond empirical risk minimization.
\newblock {\em arXiv preprint arXiv:1710.09412}, 2017.

\bibitem{zhang2018image}
Yulun Zhang, Kunpeng Li, Kai Li, Lichen Wang, Bineng Zhong, and Yun Fu.
\newblock Image super-resolution using very deep residual channel attention
  networks.
\newblock In {\em ECCV}, 2018.

\bibitem{zhang2019rnan}
Yulun Zhang, Kunpeng Li, Kai Li, Bineng Zhong, and Yun Fu.
\newblock Residual non-local attention networks for image restoration.
\newblock In {\em ICLR}, 2019.

\bibitem{zhang2018residual}
Yulun Zhang, Yapeng Tian, Yu Kong, Bineng Zhong, and Yun Fu.
\newblock Residual dense network for image super-resolution.
\newblock In {\em CVPR}, 2018.

\bibitem{Zhang2021ContextRA}
Yulun Zhang, Donglai Wei, Can Qin, Huan Wang, Hanspeter Pfister, and Yun Fu.
\newblock Context reasoning attention network for image super-resolution.
\newblock In {\em ICCV}, 2021.

\bibitem{zhou2020cross}
Shangchen Zhou, Jiawei Zhang, Wangmeng Zuo, and Chen~Change Loy.
\newblock Cross-scale internal graph neural network for image super-resolution.
\newblock In {\em NeurIPS}, 2020.

\end{thebibliography}
}

\newpage
\appendix
\section{Additional Experiments}

In Table~\red{3} we have demonstrated the (positive and negative) effects of different techniques when working with the RCAN~\cite{zhang2018image} model. This section shows an even longer training schedule, a comparison between training from scratch and warm-start, and two additional techniques we tested, which can be helpful data points for future work.

\bfsection{Even Longer Training} In Sec.~\red{4} we showed that the updated RCAN baseline is optimized for 80K iterations, and our RCAN-it is optimized for 200K iterations (160K training + 40K large-patch finetuning). RCAN-it is trained on 4 Nvidia V100 GPUs for 4 days and outperforms nearly all CNN-based approaches published in the last three years after the release of RCAN~\cite{zhang2018image}. Here we show another version called RCAN-it$\star$, which is finetuned for an additional 160K iterations using the large-patch scheme (\ie, $64\times64$), which is a total of 360K iterations. The total training time is similar to the original protocol (about 7 days). We show that the finetuning can further increase the performance on all benchmark sets (Table.~\ref{tab:supp_longer}), again demonstrating our claim that the model performance is restricted by {\em underfitting}. Specifically, RCAN-it$\star$ achieves a state-of-the-art PSNR of {\bf 40.10} for $\times2$ SR on the Manga109 dataset. 

Our results illustrate that using the large-batch scheme can effectively improve the performance of RCAN by a large margin with the same time budget. In the follow-up research, we will investigate how to improve training efficiency while keeping or improving current best results.

\bfsection{From scratch \vs warm start} In Sec.~\red{3} of the main text, we describe how we use {\em warm start}, which initialize $\times3$ and $\times4$ models with pretrained $\times2$ SR model weights to save training time. However, we are also aware of works in image classification showing that warm-starting can yield worse generalization performance than models trained from scratch~\cite{ash2020warm}. Therefore we also investigate if warm-starting influences the performance of RCAN in SR. 

\begin{table}[t]
\footnotesize
\center
\begin{center}
\caption{Quantitative comparison between updated RCAN baseline, RCAN-it, and RCAN-it$\star$. Results are evaluated by the PSNR (dB) metric for $\times2$ SR with self-ensemble. RCAN-it$\star$ improves RCAN-it with even longer large-patch finetuning.
}

\label{tab:supp_longer}
\resizebox{\columnwidth}{!}{%
\begin{tabular}{lcccccc}
\toprule
Method & Set5 & Set14 & B100 & Urban100 & Manga109
\\
\midrule
RCAN  & 38.35 & 34.42 & 32.46 & 33.61 & 39.87\\
RCAN-it & 38.41 & 34.55 & 32.51 & 33.79 & 40.04\\
RCAN-it$\star$ & {\bf 38.42} & {\bf 34.57} & {\bf 32.53} & {\bf 33.87} & {\bf 40.10}\\
\bottomrule
\end{tabular}
} 
\end{center}
\end{table}
\begin{table}[t]
\footnotesize
\center
\begin{center}
\caption{
Quantitative comparison between training from scratch and {\em warm-start}. Results are evaluated by the PSNR (dB) metric for $\times4$ SR with self-ensemble. Warm-start can save the training time by $50\%$ compared with training from scratch.
}

\label{tab:supp_scratch}
\resizebox{\columnwidth}{!}{%
\begin{tabular}{lcccccc}
\toprule
Method & Set5 & Set14 & B100 & Urban100 & Manga109
\\
\midrule
Scratch  & 32.81 & 29.08 & 27.91 & 27.32 & 31.98\\
Warm-start & 32.81 & 29.08 & 27.91 & 27.33 & 32.01\\
\bottomrule
\end{tabular}
} 
\end{center}
\end{table}
\begin{table*}[t]
\center
\begin{center}
\caption{
Additional ablation studies of model and training options. The impact of {\setlength{\fboxsep}{2pt}\textbf{\colorbox{blue!20}{architecture modifications}}} and {\setlength{\fboxsep}{2pt}\textbf{\colorbox{yellow!30}{training strategies}}} are shown upon the updated RCAN baseline (highlighted in Table~\red{2}) for $\times 2$ SR. Results are produced with self-ensemble.
}
\label{tab:additional}
\resizebox{\textwidth}{!}{%
\begin{tabular}{lcccccccccccc}
\toprule
\multirow{2}{*}{Configuration}
& \multicolumn{2}{c}{Set5} 
& \multicolumn{2}{c}{Set14} 
& \multicolumn{2}{c}{B100} 
& \multicolumn{2}{c}{Urban100} 
& \multicolumn{2}{c}{Manga109}
& \multicolumn{2}{c}{Average}
\\
\cmidrule{2-13}
& PSNR & $\Delta$ 
& PSNR & $\Delta$
& PSNR & $\Delta$
& PSNR & $\Delta$
& PSNR & $\Delta$
& PSNR & $\Delta$
\\
\midrule
Baseline &
38.350 & - &	34.415 & - & 32.463 & -	& 33.610 & - & 39.866 & - & 35.741 & - \\
\cellcolor{blue!20} w/ Deformable Conv.
& 38.153 & \textbf{\textcolor{red!60}{-0.197}}
& 33.829 & \textbf{\textcolor{red!60}{-0.586}} 
& 32.288 & \textbf{\textcolor{red!60}{-0.175}} 
& 32.632 & \textbf{\textcolor{red!60}{-0.977}} 
& 39.294 & \textbf{\textcolor{red!60}{-0.572}}
& 35.239 & \textbf{\textcolor{red!60}{-0.501}} \\
\cellcolor{yellow!30} w/ Rejection Sampling
& 38.331 & \textbf{\textcolor{red!60}{-0.019}}
& 34.371 & \textbf{\textcolor{red!60}{-0.044}} 
& 32.451 & \textbf{\textcolor{red!60}{-0.012}} 
& 33.573 & \textbf{\textcolor{red!60}{-0.037}} 
& 39.763 & \textbf{\textcolor{red!60}{-0.103}}
& 35.698 & \textbf{\textcolor{red!60}{-0.043}} \\
\bottomrule
\end{tabular}
} 
\end{center}
\end{table*}

Table~\ref{tab:supp_scratch} shows that for the $\times4$ SR task, warm-start (80K iterations) achieves better or comparable results as training from scratch  (160K iterations) while saving $50\%$ training time. This observation demonstrates that warm-start is an ecological choice for training deep SR architectures and suggests the differences between image recognition and super-resolution models. The good results of warm-start may also be because the training data for different scales are shared (with different down-sampling ratios), and we will explore that in future work.

\bfsection{Rejection Sampling} Existing SR frameworks generate training data by randomly sampling patches uniformly from all positions. However, image regions with a limited variance of shape and texture (\eg, sky) may occupy a large portion of the image and get sampled with high probability, which may not contribute much to the training. Therefore we test a rejection sampling scheme to let the model focus on relatively challenging regions. Specifically, given a sampled pair of low-resolution (LR) and high-resolution (HR) patches, we upsample the low-resolution patch with bicubic interpolation and calculate the PSNR between the upsampled patch and HR patch (please note that the LR images are also downsampled with the bicubic kernel). Higher PSNR means the patch is less challenging as bicubic sampling can already achieve a satisfactory score. Therefore we reject this pair of training patches with a probability $p$ and sample another pair instead. We calculate the statistics from 5,000 $48\times48$ input patches for the $\times2$ SR task, and notice the average PSNR score is around 24. Therefore we set the threshold to 24 and $p=0.8$ so that a sampled patch with PSNR$\geq24.0$ is rejected with a probability of 0.8. 

We show the results in Table~\ref{tab:additional}. Different from what we expected, the results are worse using the rejection sampling technique. We argue that since the problem restricting model performance is {\em underfitting} as we demonstrated in Figure~\red{2} and Table~\red{3}, changing the input distribution can have a negative influence as some relatively easy regions are still not well fitted by the RCAN model. 

\bfsection{Deformable Convolution} Besides the SiLU activation~\cite{hendrycks2016gaussian}, we also tested another architecture modification called deformable convolution~\cite{dai2017deformable} in the RCAN model. The idea of deformable convolution is to alleviate the grid structure of convolution operation to increase the field-of-view (FoV). To test if this design can help RCAN in the SR tasks, we substitute the final convolution layer in each residual group (RG) to a deformable convolution layer. The results in Table~\ref{tab:additional} show that deformable convolution decreases the baseline performance. We argue that learning the offset in the deformable convolution layers increases the difficulty in optimizing the deep SR architecture.

\section{Visual Comparison}

\begin{figure*}[t]
\scriptsize
\centering
\begin{tabular}{cc}
\hspace{-0.4cm}
\begin{adjustbox}{valign=t}
\begin{tabular}{c}
\includegraphics[width=0.197\textwidth]{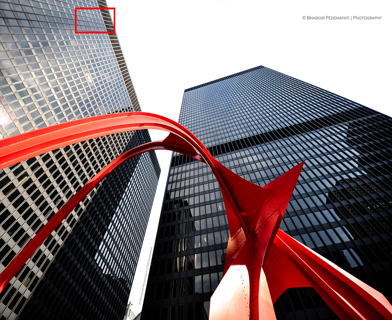}
\\
Urban100: img\_062 ($\times$4)
\end{tabular}
\end{adjustbox}
\hspace{-0.46cm}
\begin{adjustbox}{valign=t}
\begin{tabular}{cccccc}
\includegraphics[width=0.138\textwidth]{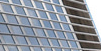} \hspace{-4mm} &
\includegraphics[width=0.138\textwidth]{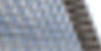} \hspace{-4mm} &
\includegraphics[width=0.138\textwidth]{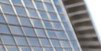} \hspace{-4mm} &
\includegraphics[width=0.138\textwidth]{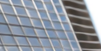} \hspace{-4mm} &
\includegraphics[width=0.138\textwidth]{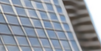} \hspace{-4mm} 
\\
HQ \hspace{-4mm} &
Bicubic \hspace{-4mm} &
EDSR~\cite{lim2017enhanced} \hspace{-4mm} &
RDN~\cite{zhang2018residual} \hspace{-4mm} &
RCAN~\cite{zhang2018image} \hspace{-4mm}
\\
\includegraphics[width=0.138\textwidth]{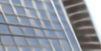} \hspace{-4mm} &
\includegraphics[width=0.138\textwidth]{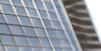} \hspace{-4mm} &
\includegraphics[width=0.138\textwidth]{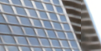} \hspace{-4mm} &
\includegraphics[width=0.138\textwidth]{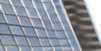} \hspace{-4mm} &
\includegraphics[width=0.138\textwidth]{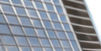} \hspace{-4mm}  
\\ 
SRFBN~\cite{li2019feedback} \hspace{-4mm} &
IGNN~\cite{zhou2020cross} \hspace{-4mm} &
CSNLN~\cite{mei2020image} \hspace{-4mm} &
RFANet~\cite{liu2020residual}  \hspace{-4mm} &
RCAN-it (ours) \hspace{-4mm}
\\
\end{tabular}
\end{adjustbox}
\vspace{1mm}
\\

\hspace{-0.4cm}
\begin{adjustbox}{valign=t}
\begin{tabular}{c}
\includegraphics[width=0.197\textwidth]{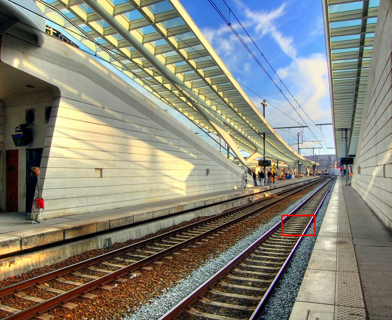}
\\
Urban100: img\_069 ($\times$4)
\end{tabular}
\end{adjustbox}
\hspace{-0.46cm}
\begin{adjustbox}{valign=t}
\begin{tabular}{cccccc}
\includegraphics[width=0.138\textwidth]{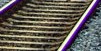} \hspace{-4mm} &
\includegraphics[width=0.138\textwidth]{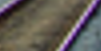} \hspace{-4mm} &
\includegraphics[width=0.138\textwidth]{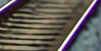} \hspace{-4mm} &
\includegraphics[width=0.138\textwidth]{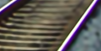} \hspace{-4mm} &
\includegraphics[width=0.138\textwidth]{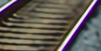} \hspace{-4mm} 
\\
HQ \hspace{-4mm} &
Bicubic \hspace{-4mm} &
EDSR~\cite{lim2017enhanced} \hspace{-4mm} &
RDN~\cite{zhang2018residual} \hspace{-4mm} &
RCAN~\cite{zhang2018image} \hspace{-4mm}
\\
\includegraphics[width=0.138\textwidth]{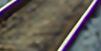} \hspace{-4mm} &
\includegraphics[width=0.138\textwidth]{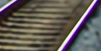} \hspace{-4mm} &
\includegraphics[width=0.138\textwidth]{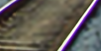} \hspace{-4mm} &
\includegraphics[width=0.138\textwidth]{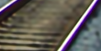} \hspace{-4mm} &
\includegraphics[width=0.138\textwidth]{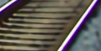} \hspace{-4mm}  
\\ 
SRFBN~\cite{li2019feedback} \hspace{-4mm} &
IGNN~\cite{zhou2020cross} \hspace{-4mm} &
CSNLN~\cite{mei2020image} \hspace{-4mm} &
RFANet~\cite{liu2020residual}  \hspace{-4mm} &
RCAN-it (ours) \hspace{-4mm}
\\
\end{tabular}
\end{adjustbox}
\vspace{1mm}
\\

\hspace{-0.4cm}
\begin{adjustbox}{valign=t}
\begin{tabular}{c}
\includegraphics[width=0.197\textwidth]{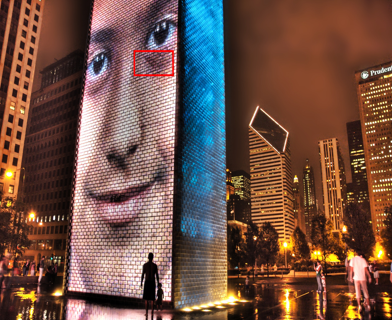}
\\
Urban100: img\_076 ($\times$4)
\end{tabular}
\end{adjustbox}
\hspace{-0.46cm}
\begin{adjustbox}{valign=t}
\begin{tabular}{cccccc}
\includegraphics[width=0.138\textwidth]{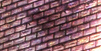} \hspace{-4mm} &
\includegraphics[width=0.138\textwidth]{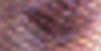} \hspace{-4mm} &
\includegraphics[width=0.138\textwidth]{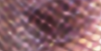} \hspace{-4mm} &
\includegraphics[width=0.138\textwidth]{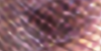} \hspace{-4mm} &
\includegraphics[width=0.138\textwidth]{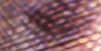} \hspace{-4mm} 
\\
HQ \hspace{-4mm} &
Bicubic \hspace{-4mm} &
EDSR~\cite{lim2017enhanced} \hspace{-4mm} &
RDN~\cite{zhang2018residual} \hspace{-4mm} &
RCAN~\cite{zhang2018image} \hspace{-4mm}
\\
\includegraphics[width=0.138\textwidth]{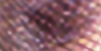} \hspace{-4mm} &
\includegraphics[width=0.138\textwidth]{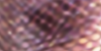} \hspace{-4mm} &
\includegraphics[width=0.138\textwidth]{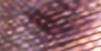} \hspace{-4mm} &
\includegraphics[width=0.138\textwidth]{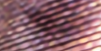} \hspace{-4mm} &
\includegraphics[width=0.138\textwidth]{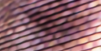} \hspace{-4mm}  
\\ 
SRFBN~\cite{li2019feedback} \hspace{-4mm} &
IGNN~\cite{zhou2020cross} \hspace{-4mm} &
CSNLN~\cite{mei2020image} \hspace{-4mm} &
RFANet~\cite{liu2020residual}  \hspace{-4mm} &
RCAN-it (ours) \hspace{-4mm}
\\
\end{tabular}
\end{adjustbox}
\vspace{1mm}
\\

\hspace{-0.4cm}
\begin{adjustbox}{valign=t}
\begin{tabular}{c}
\includegraphics[width=0.197\textwidth]{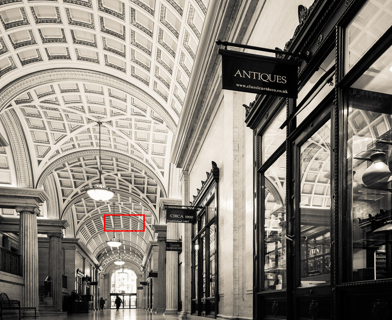}
\\
Urban100: img\_083 ($\times$4)
\end{tabular}
\end{adjustbox}
\hspace{-0.46cm}
\begin{adjustbox}{valign=t}
\begin{tabular}{cccccc}
\includegraphics[width=0.138\textwidth]{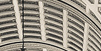} \hspace{-4mm} &
\includegraphics[width=0.138\textwidth]{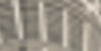} \hspace{-4mm} &
\includegraphics[width=0.138\textwidth]{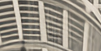} \hspace{-4mm} &
\includegraphics[width=0.138\textwidth]{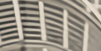} \hspace{-4mm} &
\includegraphics[width=0.138\textwidth]{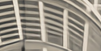} \hspace{-4mm} 
\\
HQ \hspace{-4mm} &
Bicubic \hspace{-4mm} &
EDSR~\cite{lim2017enhanced} \hspace{-4mm} &
RDN~\cite{zhang2018residual} \hspace{-4mm} &
RCAN~\cite{zhang2018image} \hspace{-4mm}
\\
\includegraphics[width=0.138\textwidth]{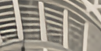} \hspace{-4mm} &
\includegraphics[width=0.138\textwidth]{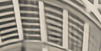} \hspace{-4mm} &
\includegraphics[width=0.138\textwidth]{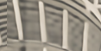} \hspace{-4mm} &
\includegraphics[width=0.138\textwidth]{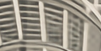} \hspace{-4mm} &
\includegraphics[width=0.138\textwidth]{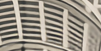} \hspace{-4mm}  
\\ 
SRFBN~\cite{li2019feedback} \hspace{-4mm} &
IGNN~\cite{zhou2020cross} \hspace{-4mm} &
CSNLN~\cite{mei2020image} \hspace{-4mm} &
RFANet~\cite{liu2020residual}  \hspace{-4mm} &
RCAN-it (ours) \hspace{-4mm}
\\
\end{tabular}
\end{adjustbox}
\vspace{1mm}
\\

\hspace{-0.4cm}
\begin{adjustbox}{valign=t}
\begin{tabular}{c}
\includegraphics[width=0.197\textwidth]{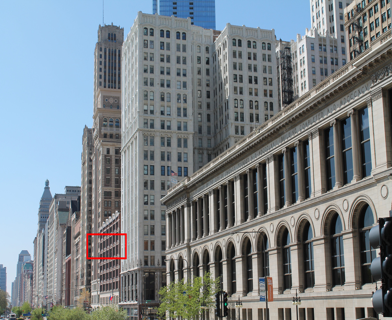}
\\
Urban100: img\_097 ($\times$4)
\end{tabular}
\end{adjustbox}
\hspace{-0.46cm}
\begin{adjustbox}{valign=t}
\begin{tabular}{cccccc}
\includegraphics[width=0.138\textwidth]{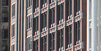} \hspace{-4mm} &
\includegraphics[width=0.138\textwidth]{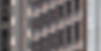} \hspace{-4mm} &
\includegraphics[width=0.138\textwidth]{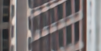} \hspace{-4mm} &
\includegraphics[width=0.138\textwidth]{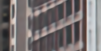} \hspace{-4mm} &
\includegraphics[width=0.138\textwidth]{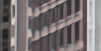} \hspace{-4mm} 
\\
HQ \hspace{-4mm} &
Bicubic \hspace{-4mm} &
EDSR~\cite{lim2017enhanced} \hspace{-4mm} &
RDN~\cite{zhang2018residual} \hspace{-4mm} &
RCAN~\cite{zhang2018image} \hspace{-4mm}
\\
\includegraphics[width=0.138\textwidth]{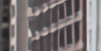} \hspace{-4mm} &
\includegraphics[width=0.138\textwidth]{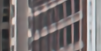} \hspace{-4mm} &
\includegraphics[width=0.138\textwidth]{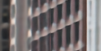} \hspace{-4mm} &
\includegraphics[width=0.138\textwidth]{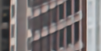} \hspace{-4mm} &
\includegraphics[width=0.138\textwidth]{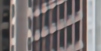} \hspace{-4mm}  
\\ 
SRFBN~\cite{li2019feedback} \hspace{-4mm} &
IGNN~\cite{zhou2020cross} \hspace{-4mm} &
CSNLN~\cite{mei2020image} \hspace{-4mm} &
RFANet~\cite{liu2020residual}  \hspace{-4mm} &
RCAN-it (ours) \hspace{-4mm}
\\
\end{tabular}
\end{adjustbox}
\vspace{1mm}
\\

\hspace{-0.4cm}
\begin{adjustbox}{valign=t}
\begin{tabular}{c}
\includegraphics[width=0.197\textwidth]{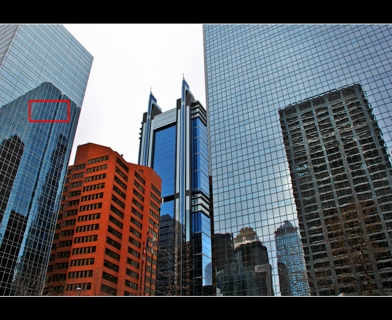}
\\
Urban100: img\_099 ($\times$4)
\end{tabular}
\end{adjustbox}
\hspace{-0.46cm}
\begin{adjustbox}{valign=t}
\begin{tabular}{cccccc}
\includegraphics[width=0.138\textwidth]{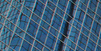} \hspace{-4mm} &
\includegraphics[width=0.138\textwidth]{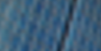} \hspace{-4mm} &
\includegraphics[width=0.138\textwidth]{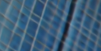} \hspace{-4mm} &
\includegraphics[width=0.138\textwidth]{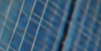} \hspace{-4mm} &
\includegraphics[width=0.138\textwidth]{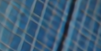} \hspace{-4mm} 
\\
HQ \hspace{-4mm} &
Bicubic \hspace{-4mm} &
EDSR~\cite{lim2017enhanced} \hspace{-4mm} &
RDN~\cite{zhang2018residual} \hspace{-4mm} &
RCAN~\cite{zhang2018image} \hspace{-4mm}
\\
\includegraphics[width=0.138\textwidth]{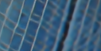} \hspace{-4mm} &
\includegraphics[width=0.138\textwidth]{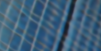} \hspace{-4mm} &
\includegraphics[width=0.138\textwidth]{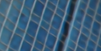} \hspace{-4mm} &
\includegraphics[width=0.138\textwidth]{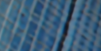} \hspace{-4mm} &
\includegraphics[width=0.138\textwidth]{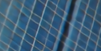} \hspace{-4mm}  
\\ 
SRFBN~\cite{li2019feedback} \hspace{-4mm} &
IGNN~\cite{zhou2020cross} \hspace{-4mm} &
CSNLN~\cite{mei2020image} \hspace{-4mm} &
RFANet~\cite{liu2020residual}  \hspace{-4mm} &
RCAN-it (ours) \hspace{-4mm}
\\
\end{tabular}
\end{adjustbox}

\end{tabular}
\caption{
Additional visual comparisons for the 4$\times$ SR task on the Urban100 dataset. Our RCAN-it (RCAN with improved training) can better reconstruct high-frequency details compared with previous approaches.
}
\label{fig:cran_visual_result_SRBIX4_supp_1}
\end{figure*}

In Figure~\red{3}, we show the visual comparisons with existing CNN-based approaches on the $\times4$ SR task. Here, we show more visual comparisons with previous SR models on the Urban100 dataset in Figure~\ref{fig:cran_visual_result_SRBIX4_supp_1}. Our RCAN-it (\ie, RCAN with improved training) can better restore the high-frequency details that are smoothed in LR images.

\section{Code}

We release the code (with MIT License) to reproduce all our experiments in this paper. Our implementation is mainly based on PyTorch~\cite{paszke2019pytorch}, also with public code from (1) EDSR\footnote{\url{https://github.com/sanghyun-son/EDSR-PyTorch}}~\cite{lim2017enhanced} official implementation (with MIT License), (2) Detectron2\footnote{\url{https://github.com/facebookresearch/detectron2}} (with Apache-2.0 License) and (3) torch-optimizer \footnote{\url{https://github.com/jettify/pytorch-optimizer}} (with Apache-2.0 License). We thank the contributors of the those libraries.
\end{document}